%% file: neurips_2026.tex
\documentclass{article}

\usepackage[numbers]{natbib}
\usepackage[main, final]{neurips_2026}
\usepackage[ruled,vlined,linesnumbered]{algorithm2e}
\usepackage{float}
\usepackage{enumerate}
\usepackage{graphicx}
\usepackage{amsmath}
\usepackage{comment} 
\usepackage{enumitem}
\setlist{nosep}
\usepackage{titletoc}
\usepackage{adjustbox}
\usepackage{booktabs}
\usepackage{float}

\usepackage[dvipsnames]{xcolor}

\usepackage[utf8]{inputenc} 
\usepackage[T1]{fontenc}    
\usepackage{hyperref}       
\usepackage{url}            
\usepackage{booktabs}       
\usepackage{amsfonts}       
\usepackage{nicefrac}       
\usepackage{microtype}      
\usepackage{xcolor}         

\title{Oracle-Budgeted Protein Design with Stochastic Search and Iterative Imitation}

\author{%
  Ashima Khanna-Reiter\textsuperscript{1}
  \qquad
  Dominik G. Grimm\textsuperscript{1,2}
  \\[0.5em]
  \textsuperscript{1}
  Technical University of Munich, TUM Campus Straubing for Biotechnology and Sustainability,\\Machine Learning for Sustainable Processes and Materials, Germany\\
  \textsuperscript{2}
  Technical University of Munich, TUM School of Computation, Information and Technology, Germany\\
  \texttt{ashima.khanna@tum.de, }
  \texttt{dominik.grimm@tum.de}
}

\begin{document}

\maketitle %

\begin{abstract}
Optimizing protein sequences under tight oracle budgets requires methods that can explore a vast combinatorial mutational search space. These methods must also remain robust in the presence of surrogate noise and propose mutations on positions where substitutions are predicted to have favorable effects. We introduce SILO (\textbf{S}elf-improvement \textbf{I}mitation \textbf{L}earning \textbf{O}ptimization), a trajectory-level approach for oracle-budgeted protein design. SILO combines a hierarchical edit policy with incremental stochastic beam search without replacement (SBS) to explore diverse candidate trajectories. A UCB-based proxy ensemble, complemented by a proxy-derived alanine-scan-inspired fitness score (AFS), guides candidate selection for \textit{in silico} oracle evaluation. The policy is then updated by next-action cross-entropy imitation on the round's best oracle-labeled trajectories. Across eight \textit{in silico} protein fitness landscapes and five strong baselines from prior work, SILO achieves competitive performance within the evaluated oracle budget, often demonstrating faster early-stage improvement than other reproduced baselines. Two-sided paired \textit{t}-tests show positive mean differences in favor of SILO across all eight landscapes for both maximum fitness and top-100 mean fitness, with statistically significant improvements after Holm correction on two and four landscapes, respectively. Under low-labeled data and noisy-proxy conditions, SILO shows competitive, task-dependent performance. Ablations identify SBS as the primary driver of performance, while AFS guidance and iterative imitation provide complementary, task-dependent benefits within the oracle-budgeted optimization loop. Code is available at: \url{https://github.com/grimmlab/SILO.git}
\end{abstract}
\section{Introduction}
Protein engineering enables the development of proteins with tailored functional properties, with applications ranging from therapeutics to enzyme engineering \cite{REISENBAUER2024102536, ndochinwa2024new, lagasse2017recent}. This engineering can be framed as a sequential decision making optimization problem under a strict evaluation budget, where the goal is to propose mutations to an initial (often wildtype) sequence that maximize a "fitness" function, quantifying a desired property such as thermostability, binding affinity, or catalytic efficiency. The problem is challenging due to the combinatorial size of the sequence space \cite{Hermes1990} and the prevalence of epistasis, which yields rugged, sparse, and highly non-convex fitness landscapes \cite{deVisser2014}. Locally plausible edits can produce highly non-additive outcomes, making efficient exploration critical. \\
A range of learning- and search-based methods address this problem. Reinforcement learning (RL) approaches optimize a generative policy via on-policy algorithms such as proximal policy optimization (PPO) \cite{schulman2017proximal} using learned reward models \cite{angermueller2019model, lee2024robust}. Self-play approaches based on AlphaZero \cite{silver2017mastering} combine policy-value networks with Monte Carlo Tree Search \cite{kocsis2006bandit}, enabling structured exploration through look-ahead search \cite{wang2023self, lin2025highplay}. Generative methods such as GFlowNets \cite{bengio2021flow} instead learn to sample diverse high-reward candidates via off-policy training \cite{jain2022biological} on existing datasets. \\
Despite these advances, sequence optimization under limited evaluation budgets faces three persistent challenges. First, value-based optimization can be unstable, sensitive to hyperparameters \cite{schulman2017proximal, henderson2018deep}, and prone to error propagation when value estimates are computed over noisy surrogate rewards \cite{andrychowicz2017hindsight}. Although off-policy generative methods improve stability \cite{jain2022biological}, they remain sensitive to the quality and coverage of training data. This can lead to distributional shifts and poor generalization to unseen regions \cite{schweighofer2021understanding}. Moreover, their performance degrade under unreliable reward estimates when evaluating out-of-distribution generated candidates \cite{kim2024improved, trabucco2021conservative}. Second, mutational proposals are typically position-agnostic, and edits at conserved positions can compromise protein function. Third, efficient exploration under strict evaluation budgets requires prioritizing high-value regions while avoiding redundant evaluations.  Together, these challenges motivate approaches that reduce reliance on noisy surrogate estimates for policy learning, promote diverse exploration of mutation trajectories, and prioritize candidates with favorable substitution effects for oracle evaluation.\\
Recent works in neural combinatorial optimization (NCO) have explored self-improvement imitation learning (SIL) \cite{luo2024self, pirnay2024self, corsini2024self} as an alternative to value-based optimization. SIL iteratively improves a policy by supervised next-action prediction on trajectories that produced the highest-quality solutions found so far, relying on strong sampling procedures to generate candidates. Crucially, because the training signal comes from the actions of oracle-labeled top performers rather than from value estimates over a noisy surrogate, SIL sidesteps the value-propagation failures that destabilize RL under approximate rewards. While effective in settings such as routing problems with cheap evaluation, adapting SIL to oracle-budgeted settings with limited or noisy feedback is non-trivial. Evaluations must be allocated carefully, and the sampling procedure must surface candidates that are both promising under the surrogate and have mutations that do not disrupt protein function. \\ \\
\noindent\textbf{Main contributions} Motivated by the challenges above, we introduce SILO, a trajectory-level protein sequence optimization approach that integrates diverse trajectory sampling, proxy-guided candidate selection, and iterative policy adaption. We evaluate SILO in an active learning setting under limited oracle budgets. Our contributions are:
\begin{enumerate}
\item \textbf{Hierarchical mutation search and trajectory-level learning.} SILO combines (i) a hierarchical edit policy that decomposes each mutation into a position choice followed by a residue choice, (ii) incremental stochastic beam search without replacement (SBS) \cite{kool2019stochastic} for diverse trajectory sampling, (iii) iterative policy adaptation from top oracle-evaluated trajectories using self-improvement imitation learning (SIL) \cite{pirnay2024self}. Candidates are prioritized for oracle evaluation using fitness scores derived from UCB-based proxy ensemble, complemented by a proxy-derived alanine-scan-inspired fitness score (AFS) \cite{kmicikiewicz2025prospero} as an auxillary signal.
\item \textbf{Sequence optimization under budgeted oracle calls. } We adapt these components into an iterative optimization process for protein design under strict oracle budgets within an active learning setup. Our ablations demonstrate that SBS drives the majority of the performance with AFS and SIL-based policy update providing additional, task-dependent adaptation.
\item \textbf{Empirical evaluation under standard and stress-test conditions.} Across eight \textit{in silico} protein fitness landscapes and five reproduced baselines, SILO demonstrates strong performance within the evaluated oracle budget, frequently achieving faster early-stage improvement than other reproduced baselines. While paired mean differences favor SILO across all eight landscapes for maximum fitness and mean fitness of top-100 sequences, Holm-adjusted two-sided paired \textit{t}-test identify statistically significant improvements on two landscapes for maximum fitness and four for top-100 mean fitness. Additional low-labeled data and noisy-proxy settings demonstrate competitive, task-dependent performance under challenging conditions.  

\end{enumerate}
\section{Problem formulation}
\label{sec:problem}
We consider protein sequence optimization over a discrete sequence space. Let $x \in \mathcal{V}^L$ denote a protein sequence of length ${L}$, where $\mathcal{V}$ is a finite vocabulary of $\vert\mathcal{V}\vert = 20$ (the standard amino acids). The objective is to maximize an oracle fitness function $O: \mathcal{V}^L \to \mathbb{R}$, measuring a desired property such as binding affinity, thermostability, or activity. We assume that oracle evaluations are expensive and only available through limited queries.
The optimization proceeds in an active-learning setting, following prior protein design works \cite{angermueller2019model,jain2022biological}. We assume access to an initial dataset $\mathcal{D}_0 = \{(x_i, O(x_i))\}_{i=1}^{M_0}$ of $M_0 = \vert\mathcal{D}_0\vert$ oracle-labeled sequences, and initialize a starting sequence as $x_{\text{start}} =$ argmax$_{x \in \mathcal{D}_0} O(x)$. At each active-learning round $n \in \{1,\ldots,N\}$, we select a batch $B_n \subset \mathcal{V}^L$ of $K$ candidate sequences for oracle evaluation, where \(K\) is the per-round oracle query budget and each candidate differs from $x_{start}$ by at most $d_{max}$ mutations. After observing their oracle fitness values, the dataset is updated as $\mathcal{D}_n = \mathcal{D}_{n-1} \cup \{(x, O(x)) : x \in B_n\}$, giving a total of at most $NK$ new oracle evaluations. Because direct oracle evaluation is limited, we train a proxy model $f_\phi$ on the currently available oracle-labeled dataset by minimizing 
\begin{equation}
\label{eq:proxy_loss}
\mathcal{L}({\phi}) = \mathbb{E}_{(x,O(x)) \sim \mathcal{D}_{n-1}} \left[(O(x) - f_\phi(x))^2\right]
\end{equation}
The proxy guides candidate selection before oracle evaluation, while final performance is always measured by the oracle $O$. The goal is to design sequences with higher fitness than the initial sequence within the budget $NK$.

\section{Active learning with SILO}
\label{sec:proposed_framework}
\begin{figure}[t]
  \centering
  \includegraphics[width=\linewidth]{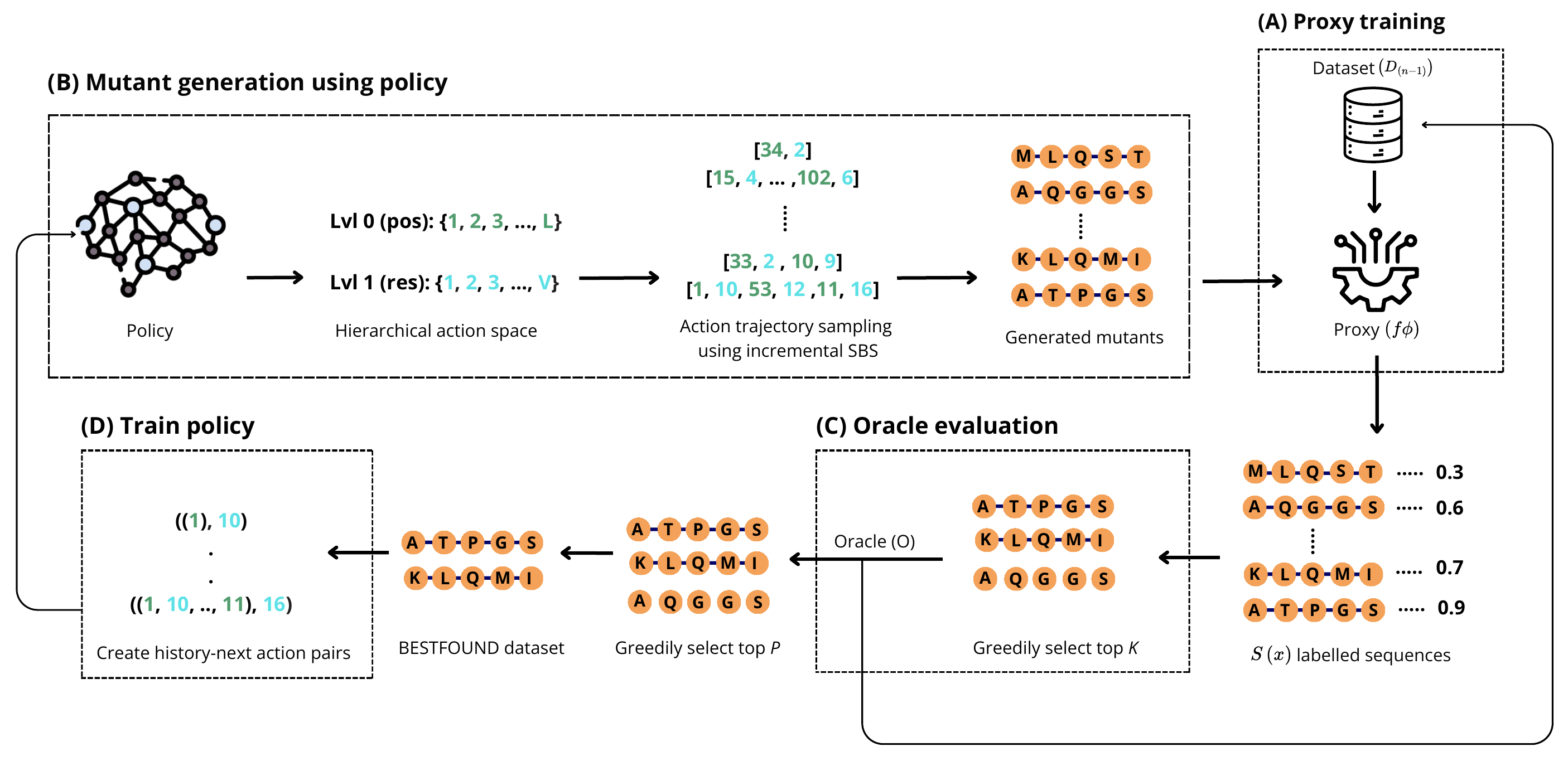}
  \caption{\textbf{Active learning with SILO. } We define the following four steps within our active learning loop: (A) A proxy ensemble model $f_\phi(x)$ is trained on the current dataset $D_{n-1}$. (B) The policy $\pi_\theta$ generates candidate mutants by sampling action trajectories (position then residue) with incremental stochastic beam search. The proxy scores the generated sequences with the objective $S(x)$ (Equation \ref{eq:combined_score}). (C) Top \textit{K} candidates are selected for oracle evaluation. (D) The top $P$ oracle-labeled sequences yield action trajectories used to train $\pi_\theta$ via next-action cross-entropy imitation. The loop repeats over $N$ active learning rounds.}
  \label{fig:silo}
\end{figure}

We follow an active learning loop, namely iterating over four steps, as illustrated in Figure \ref{fig:silo} and outlined in Algorithm \ref{alg:active}: (i) updating the proxy model $f_\phi$ by training on current dataset $D_{n-1}$ by minimizing loss as shown in Equation \ref{eq:proxy_loss}; (ii) generating mutants using policy $\pi_{\theta}$ by sampling action trajectories using incremental SBS \cite{kool2019stochastic}; (iii) evaluating the selected candidates with the oracle and appending the new sequence-score pairs to the dataset; (iv) updating the policy $\pi_{\theta}$ by training on action trajectories derived from oracle-labeled top performing sequences using a batch-wise cross entropy objective.\\
We cast candidate generation as a Markov decision process (MDP) over edit trajectories, which allows policy training via cross-entropy on action sequences rather than value estimation over generated sequences. At each round we re-select the starting sequence $x_{start}$ as the highest-oracle-scoring sequence in the current dataset $\mathcal{D}_{n-1}$, anchoring the search on the best candidate found so far. In the following subsections, we describe important components within SILO.

\begin{algorithm}[H]
\label{alg:active}
\caption{Active Learning with SILO.}
\DontPrintSemicolon
\KwIn{policy $\pi_\theta$, proxy $f_\phi$, oracle $O$, initial dataset $D_0$, oracle budget $K$, active learning rounds $N$, starting sequence $x_{start}$}
\For{$n \leftarrow 1$ \KwTo $N$}{
    Train $f_\phi$ on $D_{n-1}$: $\mathbb{E}_{(x,O(x)) \sim D_{n-1}}[(O(x) - f_\phi(x))^2]$\;
    Select starting sequence: $x_{start} \leftarrow \text{argmax}_{x\in \mathcal{D}_{n-1}} O(x)$\;
    SAMPLED $\leftarrow$ Sample mutated sequences from $\pi_\theta$ using incremental SBS with beam size $\beta$ and mutational budget $d_{max}$\;
    Score $x \in$ SAMPLED using a defined objective function $S(x)$ \;
    CANDIDATES $\leftarrow$  Select top $K$ unique sequences from SAMPLED via greedy selection\;
    Evaluate CANDIDATES with oracle $O$ and update dataset: ${D}_n \leftarrow \mathcal{D}_{n-1} \cup \{x, O(x)\}$: x $\in$ CANDIDATES\;
    BESTFOUND $\leftarrow$ top $P$ sequences from CANDIDATES ranked by $O(x)$, together with their action trajectories\;
    \For{each training step}
    {
    Sample a batch of $m$ history-next action pairs $(h^{(t)}, a^{(t)})$ uniformly from BESTFOUND trajectories\;
    Update policy $\pi_\theta$ by minimizing the cross-entropy loss:\;
    $\mathcal{L}(\theta) = - \frac{1}{m} \sum_{(h^{(t)}, a^{(t)})}$ log $\pi_\theta (a^{(t)} \mid h^{(t)})$ 
    }
}
\end{algorithm}

\subsection{Markov decision process formulation}
We formulate candidate optimization as a MDP, where an agent iteratively edits a protein sequence through a sequence of actions. A state $x^{(t)} \in \mathcal{V}^{L}$ corresponds to the edited sequence at step $t$, with $x_0 = x_{start}$ and $t \in \{0, \ldots, d_{\max}-1\}$. At each step, the agent selects an action $a^{(t)}$ from a hierarchical action space $\mathcal{A} = \mathcal{A}_0 \times \mathcal{A}_1$, sampled in two stages: 
\begin{itemize}
\item \textbf{Level-0 action} $a_0^{(t)} \in \{1, ..., L\}$ chooses a position to modify. 
\item \textbf{Level-1 action} $a_1^{(t)} \in \mathcal{V}$ chooses a replacement amino acid, conditioned on the selected position $i= a^{(t)}_0$.
\end{itemize}
The environment transitions deterministically to a new state $x^{(t + 1)}$, by substituting the amino acid at position $i$ in $x^{(t)}$ with $a_1^{(t)}$. Each episode terminates after $d_{max}$ steps, returning the final sequence $x^{(d_{max})}$, where $d_{max}$ is the per-round mutational budget defined in Section \ref{sec:problem}. This hierarchical action space decomposes each mutation into a position choice followed by an amino acid choice, providing a factorized representation for sequential mutation trajectory generation.

\subsection{Policy network architecture}
\begin{figure}[t]
  \centering
  \includegraphics[width=\linewidth]{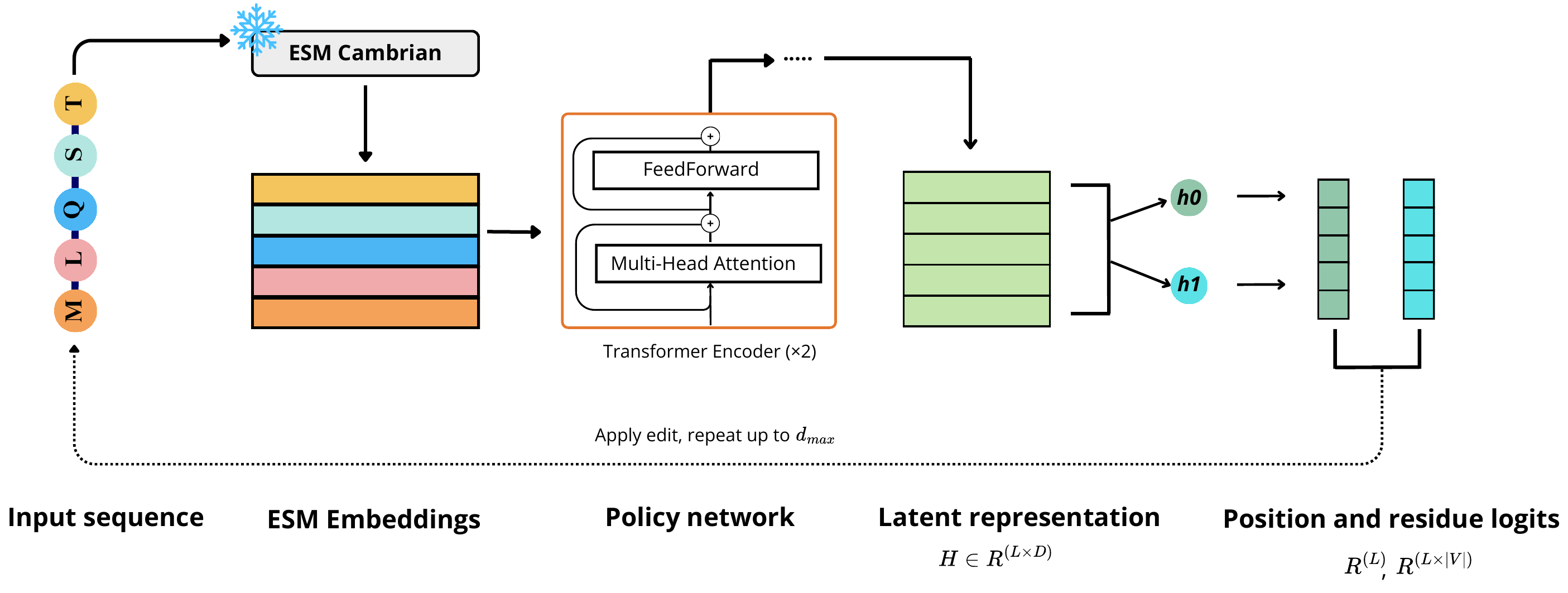}
  \caption{\textbf{Overview of the policy architecture.} An input sequence is encoded by a frozen ESM Cambrian model and the embeddings are processed by a transformer based policy. Two policy heads use latent representations from the policy to emit logits: $h_0$ over positions ($\mathbb{R}^L$) and $h_1$ over residues per position ($\mathbb{R}^{(L\times \vert\mathcal{V}\vert)}$). The policy is invoked iteratively over $d_{max}$ edit steps, updating the input sequence after each step.}
  \label{fig:policy}
\end{figure}

Following the architecture used by GraphXForm for molecular design \cite{pirnay2025graphxform}, we adapt it to the protein design space. Our policy is a lightweight transformer-based decision module (Figure \ref{fig:policy}). Given an input sequence $x$, we obtain contextualized per-residue embeddings from a frozen ESM Cambrian encoder \cite{esm2024esm}, which encodes evolutionary priors. These embeddings are then processed by a stack of transformer layers \cite{vaswani2017attention} using FlashAttention \cite{dao2022flashattention} for memory efficiency, producing a latent representation $H\in \mathbb{R}^{(L\times D)}$, where $D$ is the latent dimension.\\
Two learnable policy heads operate on $H$. A level-0 head $h_0$ produces logits over sequence positions in $\mathbb{R}^L$, and a level-1 head $h_1$ produces position-conditioned logits over the amino acid vocabulary in $\mathbb{R}^{(L\times \vert\mathcal{V}\vert)}$. The final residue distribution is obtained by selecting the row corresponding to the chosen position. This factorization reduces the joint action space of size $L\times \vert\mathcal{V}\vert$ to two decisions and naturally mirrors the hierarchical structure of the action space, enabling efficient parameter sharing while decomposing decision-making into position selection followed by amino acid selection.

\subsection{Training algorithm}
We train the policy via SIL \cite{pirnay2024self}, adapting it to the budgeted oracle setting. The key idea is to treat the policy's own highest fitness solutions as pseudo-expert demonstrations for cross-entropy imitation, without external supervision or value-function estimation. \\
Within each active learning round, the current policy $\pi_{\theta}$ samples a set of candidate trajectories of mutated sequences. Each trajectory $\tau = (a^{(0)},..., a^{(d_{max}-1)})$ corresponds to a sequence of actions that converts $x_{start}$ into a final candidate $x^{d_{max}}$. The resulting candidate sequences are scored by the proxy under the objective $S(x)$ (Equation \ref{eq:combined_score}), and the top $K$ unique sequences are evaluated by the oracle $O$. From these oracle-evaluated sequences, we select top $P$ by oracle fitness as the $\text{BESTFOUND}$ set, whose corresponding action trajectories serve as imitation targets. \\
We update $\pi_{\theta}$ on history-next action pairs $\{h^{(t)}, a^{(t)}\}$ drawn from $\text{BESTFOUND}$ trajectories, where $h^{(t)} = (a^{(0)}, \ldots, a^{(t-1)})$ denotes the sequence of past actions taken until $t-1$ step and $a^{(t)}$ denotes the corresponding next action at step $t$. The policy is then trained  by minimizing the cross-entropy loss. In practice, we optimize Equation \ref{eq:CEM} by uniformly sampling batches of history-next action pairs across $\text{BESTFOUND}$ trajectories rather than processing full trajectories. 
The improved policy is then used to generate candidates in the next round. This creates a closed-loop self-improvement imitation learning process, in which the policy progressively concentrates probability mass on regions of the sequence space yielding higher oracle fitness. Unlike within the original applications of SIL \cite{pirnay2024self}, where calling an objective function for evaluation is cheap and the full candidate pool can be re-ranked, our oracle budget motivates a proxy-scored selection step (Section \ref{sec:obj}) that interposes between trajectory sampling and oracle evaluation.    
\begin{equation}
\label{eq:CEM}
\mathcal{L}({\theta}) = - \mathbb{E}_{\tau \sim \text{BESTFOUND}} \left[\sum^{d_{max}-1}_{t=0} log \pi_{\theta} (a^{(t)} \mid h^{(t)} )\right]
\end{equation}
\subsection{Sampling}
We sample diverse candidate trajectories using incremental SBS \cite{kool2019stochastic}. SBS is a stochastic variant of beam search that draws trajectories without replacement by perturbing log-probabilities with Gumbel noise (Equation\ref{eq:gumbel} in Appendix \ref{App:SBS}). Starting from $x^{(0)} = x_{start}$, SBS expands partial action trajectories by sampling from $\pi_{\theta}$. At each step, candidate continuations are ranked by perturbed log-probabilities and the top $\beta$ trajectories are retained, forming a beam. To further improve efficiency and exploration, SIL adopts an incremental formulation of SBS, which allows sampling trajectories sequentially while dynamically updating the underlying distribution. After sampling a complete trajectory, its probability mass is removed from $\pi_{\theta}$, and the remaining probabilities are renormalized, so that subsequent trajectories are drawn from the remaining probability space. This yields a diverse set of trajectories over the action space.
\subsection{Objective function}
\label{sec:obj}
Each candidate sequence $x$ is scored using the proxy ensemble $f_\phi$, which provides a mean prediction $\mu_\phi(x)$ and an uncertainty estimate $\sigma_\phi(x)$. The first scoring term reflects the predicted fitness of $x$ itself, computed as an upper-confidence-bound (UCB) acquisition function \cite{srinivas2009gaussian}: 
\begin{equation}
\label{eq:surrogate_ucb}
\text{UCB}_{\text{proxy}}(x) = \mu_\phi(x) + \gamma_{1} \sigma_\phi(x),
\end{equation}
where $\gamma_1$ controls the exploration-exploitation factor.
The second term incorporates an alanine-scan-inspired fitness signal (AFS), a proxy-derived score motivated by experimental alanine scanning \cite{cunningham1989high, kasturi1992alanine} and originally introduced by \citet{kmicikiewicz2025prospero}. Experimental alanine scanning probes functionally relevant positions by substituting them with alanine, a neutral amino acid that perturbs side-chain interactions. Our AFS score does not rely on experimental alanine-substitution measurements but uses proxy predictions to provide an auxiliary signal of potential mutational tolerance. For a set of mutated positions $M$ in $x$ selected by the agent, we construct an alanine-substituted variant $A(x)$ by replacing all positions in $M$ with alanine, and score it with the same proxy ensemble $f_\phi$:  
\begin{equation}
\label{eq:alan_ucb}
\text{UCB}_{\text{AFS}}(x) = \mu_\phi(A(x)) + \gamma_{2} \sigma_\phi(A(x)),
\end{equation}
where $\gamma_{2}$ controls the exploration-exploitation weight on the AFS term. A high predicted fitness of the alanine-substituted variant $A(x)$ suggests that the selected positions may tolerate neutral substitutions. However, this interpretation depends on the proxy accurately predicting the fitness of alanine-substituted variants and does not constitute direct evidence of functional tolerance. While prior work \cite{kmicikiewicz2025prospero} uses AFS as a filter to identify non-disruptive substitutions, we instead use it as a complementary signal combined with the proxy UCB, allowing UCB score of alanine-substituted variants to contribute directly to candidate ranking.\\
The final candidate score is the unweighted sum of both terms:
\begin{equation}
\label{eq:combined_score}
S(x) = \, \text{UCB}_{\text{proxy}}(x) + \, \text{UCB}_{\text{AFS}}(x)
\end{equation}
After scoring all sampled candidates, we greedily select the top $K$ unique sequences for oracle evaluation.
\section{Experiments}
\label{sec:experiments}
\subsection{General setup}
\label{ssec:active_learn_workflow}
We examine the performance of SILO based on the following general setup as specified by \cite{kmicikiewicz2025prospero}, which serves as a basis of all subsequent experiments. The number of active learning rounds are $N =$ 10, and the query budget to the oracle $O$  per round is $K =$ 128. The proxy $f_\phi$  architecture is an ensemble of three one-dimensional convolutional neural networks, as specified in \cite{sinai2020adalead}. Details on training $f_\phi$ per round can be found in Appendix \ref{sssec:proxy_training}. These constraints remain consistent across all comparison partners. We set $\gamma_{1}=0.1$ and $\gamma_{2}=1$ in the objective $S(x)$ (Equation \ref{eq:combined_score}), directly adopting the corresponding UCB coefficients from ProSpero without task specific tuning. A sensitivity analysis in Table \ref{tab:varying_gamma} further examines the effect of varying these coefficients on optimization performance. At each round, we instantiate $M=5$ search instances with mutational budget $d_{max} \in \{1,2,3\}$, consistent with the low-order mutations typically done in experimental directed evolution to enable cumulative optimization. Full hyperparameters are reported in Appendix \ref{ssec:hyperparameters} 
\paragraph{Protein engineering benchmarks}
We adopt the same eight benchmark datasets and evaluation protocol as used in ProSpero (see Appendix \ref{sssec:benchmark}) \cite{kmicikiewicz2025prospero} for all methods. For AAV, we used ground-truth scores from FLEXS \cite{sinai2020adalead}, while for all other tasks we follow \citet{ren2022proximal} and use TAPE \cite{rao2019evaluating} as the \textit{in silico} oracle $O$. Following \citet{kim2024improved} and \citet{kmicikiewicz2025prospero}, we replace the experimentally-measured fitness values in $D_0$ with oracle-derived scores. This isolates optimization performance from the regression error between the \textit{in silico} oracle and the original experimental measurements, ensuring that the proxy is trained against the same target that the optimization algorithms are evaluated on. 
\paragraph{Baselines}
We compare SILO against five baselines: (i) PEX and AdaLead, which are evolutionary algorithms \cite{ren2022proximal, sinai2020adalead}; (ii) GFN-AL-$\delta$CS, which is an off-policy RL method with $\delta$ conservation parameter \cite{kim2024improved}; (iii) MLDE, a machine-learning-assisted directed evolution approach \cite{tran2024protein}; (iv) ProSpero, a biologically constrained search method based on inference-time guidance of a pre-trained generative model \cite{kmicikiewicz2025prospero}. We focus our analysis on the top baselines from the "Fitness optimization" experiment in original ProSpero publication, selected based on their overall performance. For completeness, we report both our reproduced results and the originally reported performance from ProSpero in Tables \ref{tab:max_fitness_rerun} and \ref{tab:prospero_max_fitness}, respectively.
\paragraph{Evaluation metrics}
We report four metrics, with primary emphasis on (i) \textbf{maximum fitness}, derived from the sequence with the highest fitness attained amongst all evaluated sequences, and (ii) \textbf{mean fitness}, computed as the average fitness of the top-100 evaluated sequences across all active learning rounds. We additionally report (iii) \textbf{novelty}, the average Hamming distance between top-100 oracle-evaluated sequences and the wildtype sequence, and (iv) \textbf{diversity}, the average pairwise Hamming distance within the top-100 oracle sequences. Novelty and diversity characterize the exploration profile of each method rather than its optimization quality. 

\subsection{Experiment 1: Sequence optimization task}
\label{ssec:sequence_optimization_exp}
\begin{figure}[t]
  \centering
  \includegraphics[width=\linewidth]{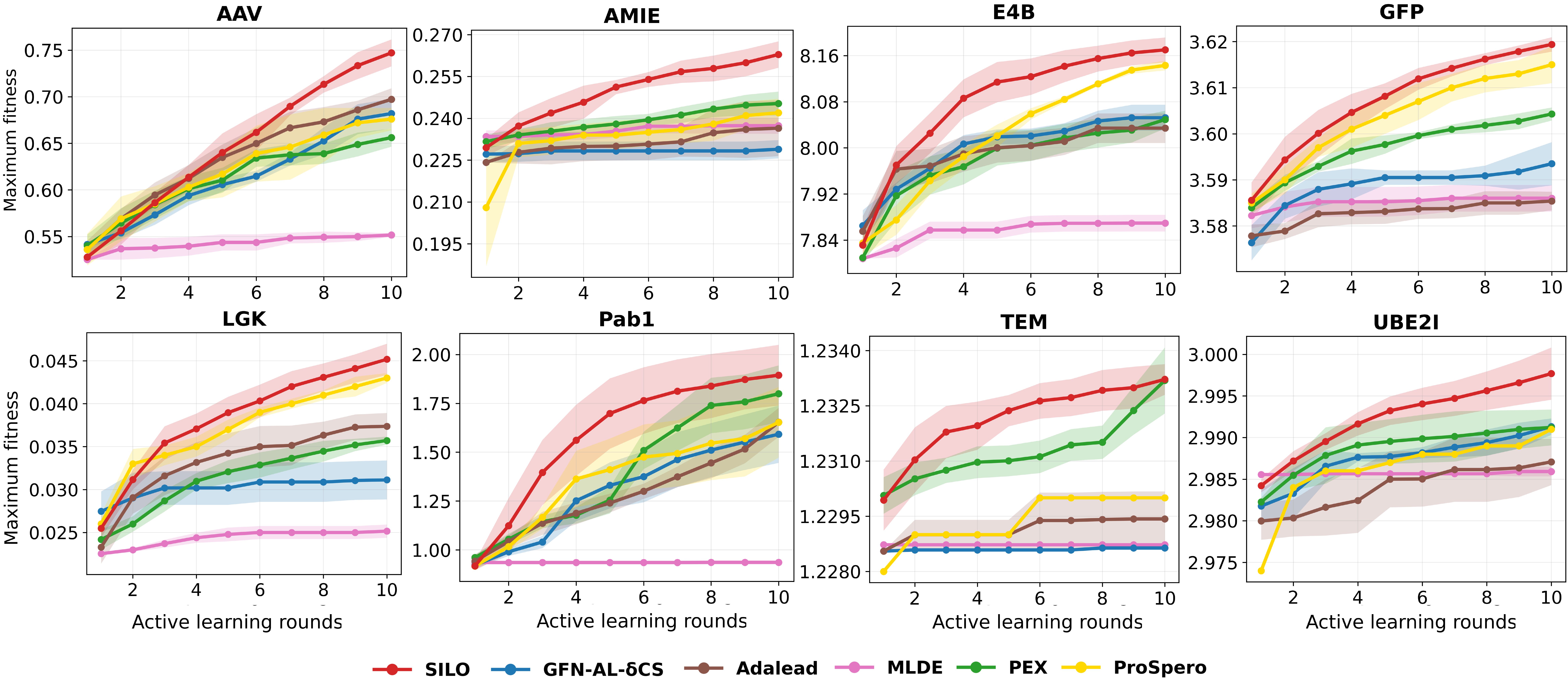}
  \caption{\textbf{Active learning performance across eight protein fitness landscapes.} Higher oracle fitness indicates better performance. All methods improve over wildtype sequence fitness $O(x_{\mathrm{WT}})$; however, SILO consistently achieves the highest maximum fitness compared to all baselines. High quality sequences are discovered in earlier rounds, indicating more efficient use of the limited oracle budgets. Shaded regions denote standard deviation over five runs.}
  \label{fig:active_learning_rounds_results}
\end{figure}

\paragraph{Setup}
We evaluate all methods under the active learning setup as described in subsection \ref{ssec:active_learn_workflow}. In this experiment, we focus on the core objective of protein fitness optimization: identifying high-fitness variants under a fixed oracle budget. We compare methods using the maximum oracle fitness and mean fitness of top-100 sequences across all rounds, measuring both peak performance and the quality of the best generated candidate set.
\paragraph{Results}  Across eight protein fitness landscapes and five reproduced baselines, SILO achieves the highest observed maximum and top-100 mean fitness on all eight landscapes within the evaluated oracle budget (Tables \ref{tab:max_fitness_rerun}, \ref{tab:mean_fitness_rerun}).  As shown in Figure \ref{fig:active_learning_rounds_results} and Table \ref{tab:max_fit_early}, SILO often discovers high-fitness sequences in earlier rounds, indicating a more efficient use of the available oracle budget. \\
To assess the statistical support of these improvements, we conduct two-sided paired \textit{t}-test over five aligned random seeds, comparing SILO against the strongest reproduced baseline selected separately for each landscape and metric. For maximum fitness, the paired differences favor SILO on all eight landscapes, with statistically significant improvements after Holm correction on AAV and AMIE (Table \ref{tab:max_paired_statistics}). UBE2I also exhibits a positive paired difference and a significant unadjusted $p$-value, but does not remain significant after correction. For top-100 mean fitness, all eight paired differences favor SILO, with statistically significant improvements on AAV, AMIE, TEM, and UBE2I after Holm correction (Table \ref{tab:mean_paired_statistics}). These findings indicate strong performance within the evaluated oracle budget, although the statistical evidence varies across landscapes and metrics. \\
We additionally evaluate sequence novelty relative to the wildtype sequence and diversity within the top-100 candidate set. SILO achieves the highest novelty on TEM and the second-highest on five other landscapes (Table~\ref{tab:novelty_rerun}). However, its top-100 candidate set exhibit lower mean diversity than other comparison partners across all eight landscapes (Table~\ref{tab:div_rerun}), indicating that gains in observed fitness do not necessarily coincide with greater within-set diversity.\\
Our primary comparisons use reproduced baselines evaluated under a consistent experimental protocol, enabling controlled, seed-aligned comparisons. We further report originally published ProSpero results in Table \ref{tab:prospero_max_fitness} under Appendix \ref{App:Baselines}, noting differences, particularly in the scores obtained for Pab1 for AdaLead.


\begin{table}[h]
\centering
\caption{Maximum fitness attained by all methods on 8 benchmark datasets across 10 active learning rounds. Reported values represent the mean $\pm$  standard deviation over 5 independent runs. Values in brackets denote wildtype $(x_{\mathrm{WT}})$ fitness. \textbf{Bold}: the best per each task. \underline{Underline}: second-best}
\label{tab:max_fitness_rerun}
\scriptsize
\setlength{\tabcolsep}{3pt}
\renewcommand{\arraystretch}{0.9}
\resizebox{\linewidth}{!}{
\begin{tabular}{lllllllll}
\toprule
Method & AAV (0.500) & AMIE (0.224) & E4B (7.743) & GFP (3.572) & LGK (0.020) & Pab1 (0.84) & TEM (1.229) & UBE2I (2.978) \\
\midrule
PEX           & 0.656 $\pm$ 0.009 & \underline{0.245 $\pm$ 0.004} & 8.049 $\pm$ 0.037 & 3.604 $\pm$ 0.037 & 0.035 $\pm$ 0.000 & \underline{1.799 $\pm$ 0.324} & \textbf{1.233 $\pm$ 0.002} & 2.991 $\pm$ 0.002 \\
AdaLead       & \underline{0.697$\pm$0.011}  & 0.236$\pm$0.009 & 8.034$\pm$0.062 & 3.585$\pm$0.002 & 0.037$\pm$0.002 & 1.653$\pm$0.165  & 1.229$\pm$0.001 & 2.987$\pm$0.002 \\
GFN-AL-$\delta$CS &  0.681$\pm$0.015 & 0.228$\pm$0.003 & 8.052$\pm$0.055 & 3.593$\pm$0.004 & 0.031$\pm$0.003  & 1.592$\pm$0.327 & 1.228$\pm$0.000 & \underline{2.991$\pm$0.000}  \\

MLDE          & 0.551 $\pm$ 0.001 &  0.237 $\pm$ 0.002 & 7.869 $\pm$ 0.035 & 3.586 $\pm$ 0.002 & 0.025 $\pm$ 0.001 & 0.936 $\pm$ 0.001 & 1.228 $\pm$ 0.000 & 2.985 $\pm$ 0.005 \\

ProSpero      & 0.676 $\pm$ 0.012 & 0.242 $\pm$ 0.005 & \underline{8.143 $\pm$ 0.021} & \underline{3.615 $\pm$ 0.004} & \underline{0.043 $\pm$ 0.001} & 1.653 $\pm$ 0.404 & \underline{1.230 $\pm$ 0.001} & 2.991 $\pm$ 0.002 \\
\midrule

SILO & {\textbf{0.747 $\pm$ 0.014}} & {\textbf{0.262 $\pm$ 0.004}} & {\textbf{8.169 $\pm$ 0.052}} & {\textbf{3.619 $\pm$ 0.002}} & {\textbf{0.045 $\pm$ 0.003}} & {\textbf{1.894 $\pm$ 0.348}} & {\textbf{1.233 $\pm$ 0.003}} & {\textbf{2.997 $\pm$ 0.003}} \\
\bottomrule
\end{tabular}
}
\end{table}

\subsection{Experiment 2: Low-labeled data regime}
\label{ssec:low_data_regime}
\begin{figure*}[t]
  \centering
  \includegraphics[width=\textwidth]{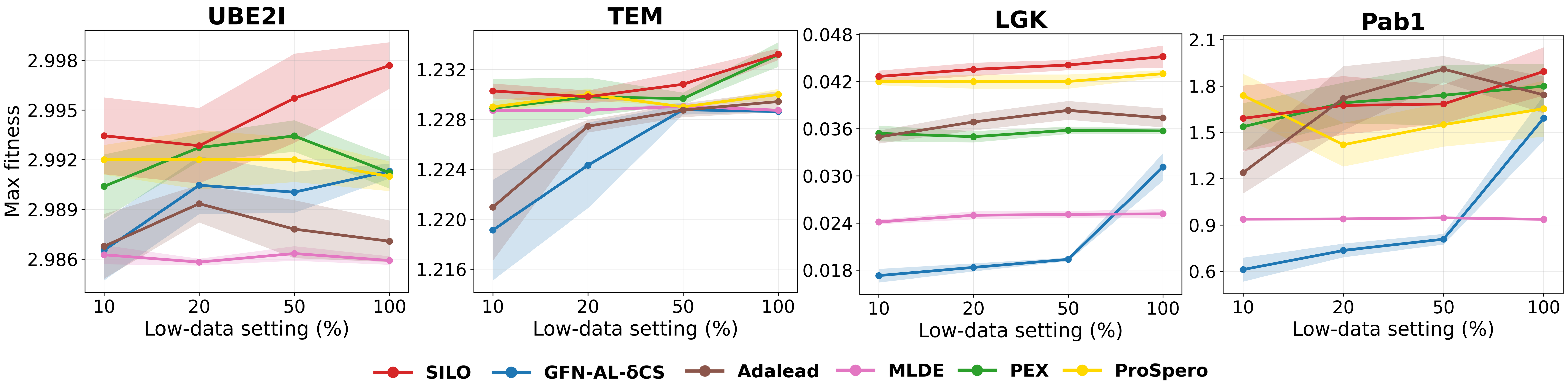}
  \caption{\textbf{Performance under limited initial training data.} Maximum fitness achieved across four protein fitness landscapes (UBE2I, TEM, LGK, and Pab1) with varying fractions of the initial training dataset ($D_0$) (10\%, 20\%, 50\%, and 100\%). SILO remains competitive across the evaluated data fractions, with relative performance varying across landscapes and data availability. }
  \label{fig:low_data}
\end{figure*}

\paragraph{Setup}
We evaluate all methods under the active learning setup as explained in Section \ref{ssec:active_learn_workflow}, but reduce the size of the initial labeled dataset $D_0$. Specifically, we randomly subsample 10\%, 20\%, and 50\% of $D_0$, using the same subsample per seed across all methods to ensure identical proxy training data. All other experimental settings are held fixed. We conduct this targeted stress test on UBE2I, TEM, LGK, and Pab1, with the goal to assess performance under reduced initial labeled data in settings where training data for proxy might be limited. We report performance with both primary and secondary metrics (Tables \ref{tab:max_mean_fitness_low_data_0.1}- \ref{tab:low_data_50_novelty_diversity}), evaluating the ability of each method to identify high-quality sequences despite reduced training data.
\paragraph{Results} 
Figure \ref{fig:low_data} compares maximum fitness across four protein fitness landscapes under varying fractions of the initial labeled dataset. SILO achieves the highest observed mean maximum fitness on LGK across all evaluated data fractions and performs competitively on UBE2I, where differences between the strongest methods remain small.  On TEM, several methods achieve comparable fitness, particularly at higher data fractions, with SILO remaining among the strongest performers. Performance on Pab1 is more variable: SILO is competitive at lower data fractions and achieves the highest observed mean fitness with the full initial dataset, while other methods lead at intermediate fractions. Overall, these results indicate that SILO maintains competitive performance under the evaluated low-data conditions, although its relative advantage depends on the protein landscape and the amount of initial labeled data.

\subsection{Experiment 3: Noisy proxy setting}
\label{ssec:noisy_proxy_setting}
\begin{figure*}[t]
  \centering
  \includegraphics[width=\textwidth]{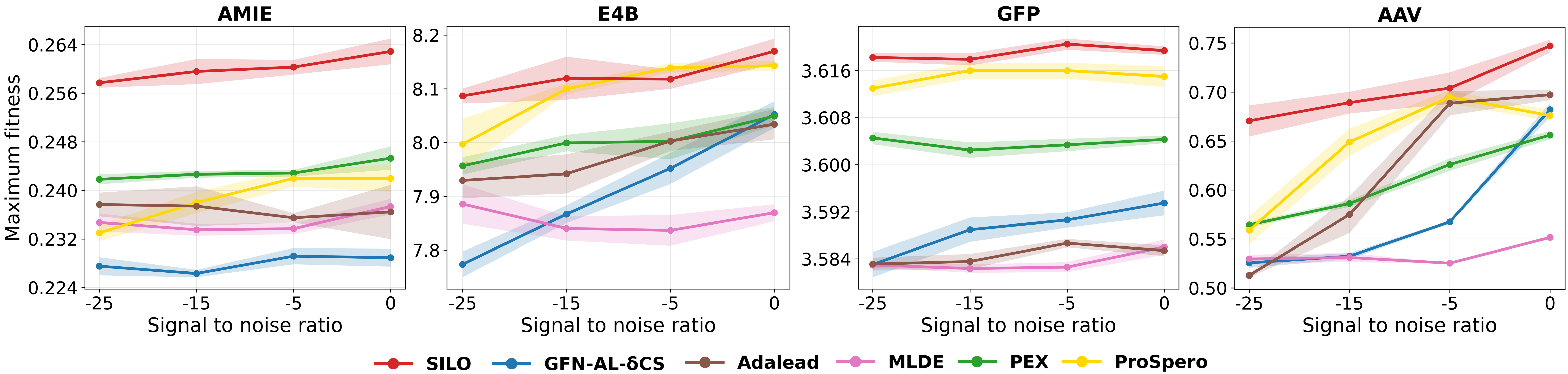}
  \caption{\textbf{Performance under noisy proxy setting.} Maximum fitness achieved across four protein fitness landscapes (AMIE, E4B, GFP, and AAV) under varying signal-to-noise ratios ($-25$, $-15$, $-5$, 0). SILO maintains competitive performance across the evaluated noise levels, achieving the highest observed mean fitness in most settings, with task-dependent differences relative to the strongest baselines.}
  \label{fig:noisy_proxy_results}
\end{figure*}

\paragraph{Setup}
In this experiment, we investigate the performance of different methods under degraded proxy quality by deliberately corrupting the proxy model, while keeping other experimental details constant. We replace the proxy $f_{\phi}$ with an ensemble of noisy oracles $f_{\epsilon}$, obtained by adding zero-mean Gaussian noise to ground-truth oracles, following the inference scheme of \cite{sinai2020adalead}. The noise magnitude is governed by the signal-to-noise ratio (SNR), with $\delta_{noise} = \sqrt{Var(D_0)} \times 10^{-SNR/10}$, where $Var(D_0)$ denotes the variance of fitness values in the initial dataset $(D_0)$. We examine three levels of surrogate noise, namely -25, -15, and -5, on AMIE, E4B, GFP and AAV. 
\paragraph{Results}
Figure \ref{fig:noisy_proxy_results} reports maximum fitness across four protein fitness landscapes under varying levels of proxy noise. Importantly, noise is introduced only into proxy predictions used for candidate ranking, while oracle evaluations remain noise-free. SILO achieves the highest observed mean fitness across all tested noise levels on AMIE, GFP, and AAV, and on three of the four noise levels on E4B, while remaining competitive with ProSpero. On AMIE and GFP, SILO's performance varies only modestly as noise increases, while on AAV and E4B, fitness decreases more noticeably under stronger noise but remains competitive with or above the strongest baselines. Several competing methods exhibit larger performance reductions under increased noise, although these trends vary across landscapes. Overall, these results suggest that SILO maintains strong performance under the evaluated synthetic proxy-noise conditions, with task-dependent sensitivity to increasing noise. However, these results do not establish robustness to noisy oracle labels, which is particularly important for SILO because policy updates are based on top oracle-evaluated trajectories.

\subsection{Experiment 4: Ablation studies}
\paragraph{Setup} We perform ablations to investigate the contributions of (i) sequence generation strategy, (ii) sampling and selection strategy, and (iii) policy learning. We first compare the fitness-guided starting-sequence selection and policy-based mutation generation against a random mutation baseline that randomly selects starting sequences, mutation positions, and amino acid substitutions. Both variants use the same Equation \ref{eq:combined_score} scoring function for candidate prioritization. Next, we compare sampling and selection strategies by replacing SBS with standard beam search (BS) and by removing the AFS term, retaining only the proxy UCB score in Equation \ref{eq:surrogate_ucb} instead of the complete scoring function in Equation \ref{eq:combined_score}. Finally, to quantify the impact of learning, we evaluate a randomly initialized frozen-policy variant, where policy parameters are not updated across active learning rounds. Table \ref{tab:ablation} reports these ablations on all eight landscapes. Complete ablation results are provided in Tables \ref{tab:ablation_complete}-\ref{tab:diversity_ablation_complete}.
\begin{table}[h]
\centering
\caption{Results of ablation studies across all eight landscapes, reporting the maximum fitness achieved over five independent runs across 10 active learning rounds. The ablations analyze the contributions of sequence generation strategy, sampling and selection strategy, and policy learning. \textbf{Bold}: the best per each task.}
\label{tab:ablation}

\setlength{\tabcolsep}{2pt}
\renewcommand{\arraystretch}{1.1}
\scriptsize

\resizebox{\linewidth}{!}{%
\begin{tabular}{lcccccccc}
\toprule
Ablation
& AAV & AMIE & E4B & GFP & LGK & Pab1 & TEM & UBE2I \\
\midrule

Random baseline
& 0.530$\pm$0.015
& 0.229$\pm$0.002
& 6.970$\pm$1.156
& 3.600$\pm$0.010
& 0.027$\pm$0.001
& 1.018$\pm$0.103
& 1.229$\pm$0.000
& 2.984$\pm$0.000 \\

SIL + BS + AFS
& 0.587$\pm$0.028
& 0.232$\pm$0.007
& 7.706$\pm$0.173
& 3.602$\pm$0.012
& 0.026$\pm$0.001
& 1.022$\pm$0.146
& 1.228$\pm$0.000
& 2.987$\pm$0.001 \\

SIL + SBS w/o AFS
& 0.694$\pm$0.019
& 0.258$\pm$0.003
& 8.140$\pm$0.062
& 3.615$\pm$0.001
& 0.042$\pm$0.001
& 1.558$\pm$0.134
& 1.232$\pm$0.000
& \textbf{2.999$\pm$0.001} \\

Frozen + SBS + AFS
& 0.706$\pm$0.013
& 0.256$\pm$0.003
& 8.149$\pm$0.041
& 3.615$\pm$0.001
& 0.042$\pm$0.001
& 1.530$\pm$0.268
& \textbf{1.233$\pm$0.002}
& 2.997$\pm$0.001 \\

\midrule

SILO
& \textbf{0.747$\pm$0.014}
& \textbf{0.262$\pm$0.004}
& \textbf{8.169$\pm$0.052}
& \textbf{3.619$\pm$0.002}
& \textbf{0.045$\pm$0.003}
& \textbf{1.894$\pm$0.348}
& \textbf{1.233$\pm$0.003}
& 2.997$\pm$0.003 \\

\bottomrule
\end{tabular}%
}
\end{table}

\paragraph{Results}
Table {\ref{tab:ablation} summarizes the ablation results across all eight protein fitness landscapes. The random mutation baseline achieves a lower maximum fitness than SILO does across all eight landscapes, despite using the same candidate-ranking objective. This suggests that the choice of starting sequences and mutation proposals contributes to optimization performance beyond the ranking function alone. Among the individual components, SBS provides the strongest and most consistent performance gains. Replacing SBS with BS, while retaining AFS and policy learning, shows statistically significant drop in performance on all eight landscapes after Holm correction (Table \ref{tab:beam_search_paired_statistics}). In contrast, AFS provides a smaller, complementary contribution: adding AFS to BS improves the observed mean maximum fitness on six landscapes and leaves it unchanged at the reported precision on the remaining two compared to the variant with BS and without AFS (Table \ref{tab:ablation_complete}). Iterative policy learning yields additional, task-dependent benefits. Compared with the frozen-policy SBS variant, full SILO achieves statistically significant improvements after Holm correction on one task, while the remaining landscapes do not reach the corrected significance threshold (Table \ref{tab:frozen_sbs_paired_statistics}). Overall, these results identify SBS as the primary driver of performance gains, with AFS and SIL based policy adaptation providing complementary benefits whose magnitude varies across tasks.
\section{Conclusion}
We presented \textbf{SILO}, a trajectory-level protein optimization approach that combines self-improvement imitation learning for policy updates, incremental stochastic beam search for diverse trajectory sampling, and proxy-guided candidate selection complemented with an alanine-scan-inspired fitness score. These components form an iterative optimization procedure that leverages top oracle-evaluated trajectories to improve the policy under limited oracle budgets.\\ 
Across eight protein benchmark tasks, SILO achieves a strong performance within the evaluated oracle budget, often exhibiting faster early-stage improvement than reproduced baselines. Paired statistical tests with Holm correction identify significant improvements on two landscapes for maximum fitness and on four landscapes for top-100 mean fitness, with positive observed paired differences across all eight landscapes for both metrics. Ablations highlight SBS as the core component for performance gains, while AFS and SIL-based policy adaptation provide complementary, task-dependent benefits. Additional experiments under low-data and noisy-proxy conditions demonstrate competitive performance across the evaluated settings.\\ 
Future work should investigate the use of SILO with larger mutation budgets, sensitivity to noisy oracle evaluations, and experimental validation beyond \textit{in silico} fitness oracles.

\section*{Funding and acknowledgments}
This work was supported by the Pythagoras Project (Grant No.~031B1407B) within the program “Climate-neutral products through biotechnology–CO$_2$ and C1 compounds as sustainable raw materials for the industrial bioeconomy (CO2BioTech)”, which is funded by the Bundesministerium für Forschung, Technologie und Raumfahrt (BMFTR). The Deutsche Forschungsgemeinschaft (DFG, German Research Foundation) further supported this work through Project No.~466387255 within the Priority Programme “SPP 2331: Machine Learning in Chemical Engineering”. The authors gratefully acknowledge the Competence Center for Digital Agriculture (KoDA) at the University of Applied Sciences Weihenstephan-Triesdorf for providing computational resources.
{
\small
\bibliographystyle{plainnat}
\bibliography{references}
\clearpage
\appendix
\startcontents[appendix]
\section*{Appendix}
\printcontents[appendix]{}{1}{\setcounter{tocdepth}{2}}
\clearpage
\section{Related work}
\paragraph{Search and surrogate-guided generative methods} 
Adapt-with-the-Leader \textbf{(AdaLead)} is a model-guided method that iteratively refines sequences by greedily applying mutations informed by a learned proxy, following the evolutionary and follow the perturbed leader approaches \cite{sinai2020adalead}. Proximal Exploration (\textbf{PEX)}, introduced by Ren et al., emphasizes conservative exploration by prioritizing lower-order mutants and close to the wildtype sequence, framing the problem of local search as proximal optimization \cite{ren2022proximal}. Within Machine Learning-Driven Evolution \textbf{(MLDE)} framework, Tran and Hy leveraged ESM-2 protein language model to generate mutants through random and importance-based masking of the wild-type sequence \cite{tran2024protein}. Lastly, \textbf{ProSpero} integrates proxy guidance with inference-time control of pretrained generative model, EvoDiff, using targeted masking and biologically constrained sampling to generate candidates without updating any learned parameters of EvoDiff across rounds \cite{kmicikiewicz2025prospero}.
\paragraph{Reinforcement learning} \textbf{DyNA-PPO} \cite{angermueller2019model} applies model-based RL to generate sequences using PPO \cite{schulman2017proximal} to guide the policy and using a learned proxy for generating high-fitness candidates. \textbf{LatProtRL} extends this paradigm by performing optimization in the latent space of ESM-2 model. A PPO-based policy is trained to navigate this latent space, enabling smoother exploration \cite{lee2024robust}. \textbf{EvoPlay} \cite{wang2023self} adopts a self-play RL framework inspired by AlphaZero \cite{silver2017mastering}, combining a policy-value network with Monte Carlo Tree Search \cite{kocsis2006bandit} to iteratively refine candidate sequences.
\paragraph{Generative and distribution-based models} Methods like GFlowNets \cite{bengio2021flow}, learn a stochastic policy by combining offline datasets with on-policy samples during training to generate high fitness candidates \cite{jain2022biological}. While they improve stability over methods like DyNA-PPO, they can be sensitive to proxy misspecification \cite{kim2024improved, trabucco2021conservative}. Building on this, Kim et al. introduced GFN-AL-$\delta$CS, which utilizes conservativeness parameter $\delta$ to balance novelty and robustness \cite{kim2024improved}. Authors of DbAS iteratively refine a generative model, i.e. variational autoencoder, to high fitness sequences under a surrogate model, effectively concentrating probability mass on promising regions \cite{brookes2018design}. CbAS extends DbAS by adding distributional regularization to mitigate surrogate bias and improve stability \cite{brookes2019conditioning}. 
\paragraph{Difference between SILO and related methods} SILO differs from above methods in several key ways. First, unlike DbAS/CbAS and GFlowNets/GFN-AL-$ \delta$CS, SILO operates in action trajectory space rather than sequence space, decomposing the problem into finer-grained decisions, rather than modeling the sequence as a whole, enabling more structured and targeted exploration of the mutational space. Second, while both SILO and ProSpero use an alanine-scan inspired score to provide feedback on disruptive positions, ProSpero relies on a pretrained generative model that is never updated after deployment. In contrast, SILO learns a reusable editing policy from top oracle-evaluated trajectories, accumulating task-specific knowledge about which positions and substitutions lead to fitness gains across rounds. Finally, unlike EvoPlay and other PPO-based approaches that employ a critic, SILO requires neither a value function nor a critic network. Instead, the policy is trained via cross-entropy loss on action trajectories of the highest-fitness oracle-evaluated mutants from each round, without directly relying on proxy-predicted fitness as a training target.
\paragraph{SILO vs current applications of SIL} The SIL paradigm from \citet{pirnay2024self}, and its applications to routing problems \cite{pirnay2024self} and molecular design \cite{pirnay2025graphxform}, trains a policy via supervised cross-entropy on action trajectories that produced the best discovered solutions. Importantly, both settings assume cheap and unlimited evaluations from the objective function. Our setting, however, imposes a strict oracle budget per round, making effective candidate prioritization essential and motivating the combination of proxy-based acquisition with an auxiliary alanine-scan-inspired fitness signal in SILO.
\section{Experiment setting and implementation details}
\label{sec:supplementary_experiments}
\subsection {Benchmark datasets}
\label{sssec:benchmark}
SILO was evaluated on eight diverse benchmark protein fitness landscapes. As mentioned in \cite{kmicikiewicz2025prospero}, the AAV and GFP benchmarks were originally introduced by Kim et al. (Apache-2.0 license) \cite{kim2024improved}, while the remaining datasets were compiled by Tran and Hy \cite{tran2024protein} (GPL-3.0 license). Oracles are taken from FLEXS \cite{sinai2020adalead} for AAV and GFP (Apache-2.0 license), and from Ren et al. \cite{ren2022proximal} for the other tasks (Apache-2.0 license).
\begin{enumerate}
\item \textbf{Adeno-associated Virus (AAV):} The task focuses on improving binding affinity of an amino
acid segment (position 450-540) of the VP1 protein located in the capsid of the Adeno-associated virus. The dataset $D_0$ containing 15,307 sequences, was generated by \citet{kim2024improved} by randomly mutating the wild-type sequence while filtering out sequences that have higher scores than the wild-type. Here, the sequence length $L =$ 90, corresponding to search space of $20^{90}$. The starting fitness is the wildtype sequence is $O(x_{WT}) = 0.500$ and its average Hamming distance to sequences in $D_0$ is Novelty ($D_0$, $x_{WT}$) = 5.05. 
\item \textbf{Aliphatic Amide Hydrolase (AMIE):} The task aims to optimize amidase sequences for increased enzymatic activity \cite{wrenbeck2017single}. Initial dataset $D_0$ contains 6417 sequences with single mutations to model the fitness landscape. The length of sequence $L =$ 341. The starting fitness is $O(x_{WT}) = 0.224$, and Novelty ($D_0$, $x_{WT}$) = 2. 
\item \textbf{Ubiquitination Factor Ube4b (E4B):} The goal is generate sequences that enhance E4B ubiquitination enzyme. \citet{starita2013activity} measured the rates of ubiquitination of the mutants to the target protein. The full dataset includes 91,032 sequences of length $ L=102$, from which \cite{kmicikiewicz2025prospero} randomly selected 10,000 sequences to form $D_0$. The starting sequence has $O(x_{WT}) =$ 7.743 and Novelty ($D_0$, $x_{WTt}$) = 5.42.
\item \textbf{Green Fluorescent Protein (GFP):} The task seeks to generate sequences with high log-fluorescence intensity \cite{sarkisyan2016local}. The $D_0$ dataset contains 10,200 sequences generated by \cite{kim2024improved}, similarly to AAV. The sequence length is $L=$238, with $O(x_{WT}) =$ 3.572 and Novelty ($D_0$, $x_{WT}$) = 42.87.
\item \textbf{Levoglucosan Kinase (LGK):} This task aims to improve enzymatic activity of \textit{levoglucosan kinase}, which converts LG to the glycolytic intermediate glucose-6-phosphate \cite{klesmith2017trade}. The initial dataset $D_0$ includes 7,633 sequences of length $L=$ 439, with $O(x_{WT}) =$ 0.020 and Novelty ($D_0$, $x_{WT}$) = 2.0.  
\item \textbf{Poly(A)-binding Protein (Pab1):} The poly(A)-binding protein Pab1 binds to polyadenosine (poly-A) sequences via its RNA recognition motif (RRM). \citet{melamed2013deep} conducted a high-throughput screening assay to measure binding fitness for approximately 36,000 double mutants of Pab1 within the RRM region. This task focuses on improving binding affinity within the RRM. As with E4B, \cite{kim2024improved} subsampled 10,000 sequences to construct $D_0$. The length of the sequence is $L = $ 75, the starting sequence has $O(x_{WT}) = $ 0.843 and Novelty ($D_0$, $x_{WT}$) = 3.95.
\item \textbf{TEM-1 $\beta$-Lactamase (TEM):} TEM-1 $\beta$-Lactamase resistance to penicillin antibiotics in \textit{E.coli }is widely studied to understand mutational effect and fitness landscape \cite{jacquier2013capturing}, \cite{bershtein2006robustness}. The aim of this task is to generate TEM variants with improved thermodynamic stability. The dataset $D_0$ derived \cite{firnberg2014comprehensive} from contains 5,199 sequences of length $L = $286. The starting sequence has fitness $O(x_{WT}) = $ 1.229 and Novelty ($D_0$, $x_{WT}$) = 2.0.
\item \textbf{SUMO E2 Conjugase (UBE2I):} Variants of the disease-relevant protein, human SUMO E2
conjugase were generated by \citet{weile2017framework} and the goal of this task is to optimize these variants for functional mapping applications. $D_0$ comprise of 3,022 sequences of length $L = $159. The starting sequence has $O(x_{WT}) = $2.978 and Novelty ($D_0$, $x_{WT}$) = 2.0.
\end{enumerate}

\subsection {Proxy architecture and training} \label{sssec:proxy_training}
For the proxy model, we use an ensemble of one-dimensional convolutional neural network \cite{kmicikiewicz2025prospero, sinai2020adalead}. The ensemble size is assigned to 3, and the model is trained using the Adam optimizer \cite{kingma2014adam} with learning rate and L2 penalty set to 0.0001. We do a random 90\%-10\% train-validation data split. The batch size is set to 256. We train the ensemble for 3000 epoches. To mitigate overfitting, we use early stopping based on a validation set comprising 10\% of the data, stopping training if the validation loss does not improve for ten consecutive evaluations.

\subsection {Baselines}
\label{App:Baselines}
We reproduce the top five comparison partners from the "Fitness optimization" experiment in ProSpero by rerunning them under the same experimental conditions, including identical hyperparameter configurations and seeds as specified in the original work \cite{kmicikiewicz2025prospero}. All methods are evaluated under a consistent experimental protocol, including the same datasets, oracle implementation, query budget, and proxy architecture. Despite these efforts, some reproduced results differ from those originally reported. We therefore use the reproduced results for our primary comparisons and report the originally published results separately for reference (Tables \ref{tab:prospero_max_fitness}, \ref{tab:prospero_mean_fitness}). All comparison partners are operated under active learning rounds $N = $ 10, number of queries to the oracle per round $K = $ 128, and the same proxy architecture.
\begin{enumerate}
\item \textbf{AdaLead \cite{sinai2020adalead}:} We employed the open-source implementations provided by Sinai et al. \cite{sinai2020adalead} available at \url{https://github.com/samsinai/FLEXS/tree/master} under the Apache-2.0 license. We used the default hyperparameters of the model, with a recombination rate of 0.2, a mutation rate of 1/L, where L is the sequence length, and a threshold $\tau = $ 0.05. The numbers of model predictions is set to 2000. 
\item \textbf{GFN-AL-$\delta$CS \cite{kim2024improved}:} For the conservative strategy GFN-AL-$\delta$CS proposed by Kim et al. \cite{kim2024improved}, we employ an adaptive $\delta$ with a maximum masking radius of 0.05 and rank-based proxy training with a reweighting factor k=0.01. We set the scaling factor $\lambda = $0.1 for AAV, E4B, and Pab1, and $\lambda = $1 for GFP, AMIE, TEM, UBE2I, and LGK. We use the publically released codebase from \url{https://github.com/hyeonahkimm/delta_cs.git} under the Apache-2.0 license. 
\item \textbf{MLDE \cite{tran2024protein}:} We use the MLDE implementation from Tran and Hy \cite{tran2024protein}, from their publicly available codebase \url{https://github.com/HySonLab/Directed_Evolution} under the GPL-3.0 license. 
Adapting their setup to perform active learning, we perform 10 rounds of surrogate-guided optimization with a population size of 128 and a beam size of 4. The masking strategy employs a random-to-importance ratio of 0.6:0.4, and unmasking is performed using ESM-2 \cite{lin2022language} with 35M parameters.
\item \textbf{PEX \cite{ren2022proximal}:} We implement PEX from Ren et al. \cite{ren2022proximal} using the official codebase \url{https://github.com/HeliXonProtein/proximal-exploration/tree/main} under the Apache-2.0 license. We use the default configuration, including 2 random mutations and a frontier neighborhood size of 5.
\item \textbf{ProSpero \cite{kmicikiewicz2025prospero}:} We use ProSpero under the official codebase \url{https://github.com/szczurek-lab/ProSpero.git} released by Kmicikiewicz, Fortuin, and Szczurek \cite{kmicikiewicz2025prospero} under the GPL-3.0 license. We use the default hyperparameters of the method, keeping the range of number of corruptions introduced within the sequences between 3-10 for shorter sequences (E4B, Pab1, AAV) and 5-15 for longer sequences (GFP, LGK, AMIE, TEM, UBE2I). Number of scans (S) for targeted masking is set to 16 and sequential monte carlo (SMC) batch size is 256. The UCB exploitation-exploration hyperparamaters for targeted masking and biologically-constrained SMC are $k = $ 1.0 and 0.1, respectively.  Additionally, we run experiments under fully deterministic setting to reinforce reproducibility. 
\end{enumerate}
\subsection {Protein sequence representations}
To compute sequence embeddings, we use a frozen ESM Cambrian \cite{esm2024esm}, a pretrained protein language model with 300M parameters, available under Cambrian Open License Agreement.
\subsection {Hyperparameters for SILO}
\label{ssec:hyperparameters}
At each round, we instantiate $M=5$ mutation-constrained search instance with the budget of 1–3 mutations and apply SBS to sample diverse trajectories. The beam size $\beta$ is 32 and the total sequences generated for proxy evaluation per round are 480. We use $P =$ 50 top oracle-evaluated candidates per round for training the policy. The batch size for training is 16. The number of transformer blocks for the policy are 2, number of heads are 16, and latent dimension $R_D$ is 512. We train the policy using Adam optimizer \cite{kingma2014adam} with a learning rate of 0.0001.  
\subsection {Evaluation metrics}
Considering $\mathcal{D}_{\text{top}} = \{ (x^{(i)}, O(x^{(i)})) \}_{i=1}^{100}$ contains the set of the top-100 oracle evaluated sequences identified across $N$ active learning rounds. We evaluate all methods using the following metrics:
\begin{enumerate}
\item \textbf{Maximum fitness:} The highest oracle fitness attained among the evaluated sequences, measuring the ability to identify high-performing candidates:
\begin{equation}
\label{eq:max_fitn_eq}
\text{MaxFitness}(\mathcal{D}_{\text{top}}) = \max_{x \in \mathcal{D}_{\text{top}}} O(x)
\end{equation}

\item \textbf{Mean fitness:} The average oracle fitness of the top-100 evaluated sequences, characterizing the quality of the highest-scoring candidate set:
\begin{equation}
\label{eq:mean_fitn_eq}
\text{MeanFitness}(\mathcal{D}_{\text{top}}) = \frac{1}{|\mathcal{D}_{\text{top}}|} \sum_{x \in \mathcal{D}_{\text{top}}} O(x)
\end{equation}

\item \textbf{Novelty:} Quantifies the average Hamming distance between the top sequences and the wildtype sequence:
\begin{equation}
\label{eq:nove_fitn_eq}
\text{Novelty}(\mathcal{D}_{\text{top}}, x_{WT}) = \frac{1}{|\mathcal{D}_{\text{top}}|} \sum_{x \in \mathcal{D}_{\text{top}}} d(x, x_{WT})
\end{equation}

\item \textbf{Diversity:} The average pairwise Hamming distance among the top-100 sequences, characterizing sequence variation within the selected set: 
\begin{equation}
\label{eq:diversity_fitn_eq}
\text{Diversity}(\mathcal{D}_{\text{top}}) = \frac{1}{|\mathcal{D}_{\text{top}}|(|\mathcal{D}_{\text{top}}| - 1)} \sum_{x, x' \in \mathcal{D}_{\text{top}, x \neq x'}} d(x, x')
\end{equation}

\end{enumerate}
\section {Discussion}
\subsection {Hardware and runtime comparisons}
\label{ssec:hardware}
Our code is developed in PyTorch \cite{paszke2019pytorch} v.2.8.0. Experiments were conducted on an NVIDIA A40 GPU using CUDA 12.2. We measured the wall-clock runtime for one complete experiment consisting of ten active learning rounds across all eight landscapes for runtime comparison. Under this setup, SILO requires $4.5$ hours. It is faster than ProSpero ($20.86$ hours) and GFN-AL-$\delta$CS ($5.66$ hours), but slower than MLDE ($1.96$ hours), PEX ($0.70$ hours), and AdaLead ($0.67$ hours). SILO uses additional sampling and policy-update computation in exchange for stronger oracle-budgeted optimization, not to minimize local computation. Thus, SILO has intermediate computational cost rather than a uniform runtime advantage. 

\begin{table}[H]
\centering
\caption{Compute time required for one complete experiment for each method. A complete experiment consists of 10 active learning rounds performed on all eight benchmark landscapes.}
\label{tab:compute_time}
\scriptsize
\begin{tabular}{lc}
\hline
\textbf{Method} & \textbf{Runtime per experiment (hours)} \\
\hline
PEX &  0.70 \\
AdaLead & 0.67 \\
GFN-AL-$\delta$CS & 5.66 \\
MLDE & 1.96 \\
ProSpero (enforcing reproducibility setting) & 20.86 \\
\midrule
SILO & 4.5 \\
\hline
\end{tabular}
\end{table}
    
\subsection {Limitations}
\label{ssec:limitations}
\textbf{Local search constraints.} Because improvements are made incrementally within a budget of 1-3 edits per round, SILO may struggle to discover distant high-fitness regions that are not reachable through a sequence of local improvements. Evaluating larger mutational budgets remains an important direction for future work. \\
\textbf{Unexamined sensitivity to oracle noise.} Our noisy-proxy experiments perturb the proxy predictions used for candidate ranking while retaining noise-free oracle evaluations. These experiments do not establish robustness to noisy oracle labels, which could affect the selection of high-performing trajectories used for imitation-based policy updates and therefore should be examined further.\\
\textbf{\textit{In silico} oracles.} All experiments use \textit{in silico} oracles rather than wet-lab measurements. Performance under real experimental conditions may differ due to batch effects, assay noise, and domain shift between the oracle model and actual biological fitness. Evaluating SILO in a wet-lab-in-the-loop setting would be an important next step toward validating practical applicability.
\subsection {Broader impact}
\label{ssec:broader_impact}
Our work improves methods for protein sequence optimization, with applications in areas such as enzyme design, biotechnology, and drug discovery. By enabling efficient search for high fitness variants under limited and noisy proxy settings, it may help accelerate the development of beneficial biological systems by reducing the burden of extensive wet lab experimentation. However, protein design methods can carry dual-use risks if misapplied. Notably, one of our benchmark tasks involves AAV caspid optimization, a protein family directly relevant to gene therapy vectors. More broadly, such optimization methods could be applied to optimize proteins with pathogenic, toxic, or otherwise harmful phenotypes. Therefore, appropriate safeguards for responsible use of these methods in downstream applications should be enforced. Furthermore, we will not release any task-specific fine-tuned policy weights for sequences flagged by biosecurity screening tools.
\section {Algorithmic details}
\subsection {Incremental stochastic beam search}
\label{App:SBS}
We employ SBS, originally introduced by Kool et al. \cite{kool2019stochastic}, as the core sampling procedure, which is a key component within SIL. SBS is a stochastic variant of beam search, which enables sampling trajectories \textbf{without replacement} by perturbing policy log-probabilities with Gumbel noise. Concretely, given a partial trajectory $a_{0:t}$, each candidate extension $a_{0:t+1}$ is scored using a perturbed log-probability and top-$\beta$ candidates are retained in the beam. 
\begin{equation}
\label{eq:gumbel}
\tilde{G}(a_{0:t+1}) = \log \pi_\theta(a_{0:t+1}) + \epsilon, 
\quad \epsilon \sim \mathrm{Gumbel}(0, 1)
\end{equation}
In its incremental formulation, SBS allows sampling multiple complete trajectories sequentially while dynamically updating the underlying distribution. After a trajectory is sampled, its probability mass is removed and the distribution is renormalized, ensuring that subsequent samples are drawn from the remaining probability space. In our framework, SBS operates over the action space, where each node in the search tree represents a partial mutation trajectory. Combined with the SIL loop, this results in a set of high-quality and diverse action trajectories used for training.
\section {Complete results}
\subsection {Sequence optimization task}
\begin{table}[H]
\centering
\caption{Mean fitness of top-100 sequences attained by all methods on 8 benchmark datasets across 10 active learning rounds. Reported values are the mean $\pm$ standard deviation over 5 independent runs. Values in brackets denote wildtype $(x_{\mathrm{WT}})$ fitness. \textbf{Bold}: the best per each task. \underline{Underline}: second-best}
\label{tab:mean_fitness_rerun}
\scriptsize
\setlength{\tabcolsep}{3pt}
\renewcommand{\arraystretch}{0.9}
\resizebox{\linewidth}{!}{
\begin{tabular}{lllllllll}
\toprule
Method & AAV (0.500) & AMIE (0.224) & E4B (7.743) & GFP (3.572) & LGK (0.020) & Pab1 (0.84) & TEM (1.229) & UBE2I (2.978) \\
\midrule
PEX           & 0.603$\pm$0.010 & \underline{0.234 $\pm$ 0.001} & 7.877$\pm$ 0.047 & 3.597 $\pm$ 0.001 & 0.033 $\pm$ 0.001 & 1.476 $\pm$ 0.224 & \underline{1.228 $\pm$ 0.000} & \underline{2.987 $\pm$ 0.002} \\

AdaLead       & \underline{0.661 $\pm$ 0.013}  & 0.228 $\pm$ 0.010 & 7.874 $\pm$ 0.064 & 3.571 $\pm$0.002 & 0.034 $\pm$ 0.003 & 1.455 $\pm$ 0.154 & 1.209$\pm$0.000 & 2.974 $\pm$ 0.002 \\

GFN-AL-$\delta$CS & 0.638$\pm$0.007 & 0.184$\pm$0.010 & 7.917$\pm$0.054 & 3.576$\pm$0.001 & 0.020$\pm$0.002 & 1.485$\pm$0.324 & 1.207$\pm$0.003 & 2.981$\pm$0.002 \\

MLDE          & 0.533$\pm$0.005 & 0.230$\pm$0.002 & 7.769$\pm$0.010 & 3.579$\pm$0.001 & 0.022$\pm$0.001 & 0.902$\pm$0.019 & 1.227$\pm$0.000 & 2.981$\pm$0.000  \\

ProSpero & 
0.632 $\pm$ 0.023 & 0.232 $\pm$ 0.005 & \underline{8.061 $\pm$ 0.010} & \underline{3.611 $\pm$ 0.003} & \underline{0.041 $\pm$ 0.001} & \underline{1.512 $\pm$ 0.425} & 1.193 $\pm$ 0.011 & 2.985 $\pm$ 0.002 \\

\midrule
SILO & {\textbf{0.727 $\pm$ 0.014}} & {\textbf{0.259 $\pm$ 0.005}} & {\textbf{8.148 $\pm$ 0.058}} & {\textbf{3.617 $\pm$ 0.001}} & {\textbf{0.044 $\pm$ 0.002}} & {\textbf{1.856 $\pm$ 0.355}} & {\textbf{1.232 $\pm$ 0.001}} & {\textbf{2.996 $\pm$ 0.002}} \\
\bottomrule
\end{tabular}
}
\end{table}

\begin{table}[H]
\centering
\caption{Average novelty of top-100 sequences generated by all methods on 8 benchmark datasets across 10 active learning rounds. Reported values are the mean $\pm$ standard deviation over 5 independent runs. \textbf{Bold}: the best per each task. \underline{Underline}: second-best}
\label{tab:novelty_rerun}
\footnotesize
\setlength{\tabcolsep}{3pt}
\renewcommand{\arraystretch}{0.9}
\resizebox{\linewidth}{!}{
\begin{tabular}{lcccccccc}
\toprule
Method & AAV & AMIE & E4B & GFP & LGK & Pab1 & TEM & UBE2I \\
\midrule
PEX & 6.694 $\pm$ 0.758 & 4.640 $\pm$ 0.799 & 3.584 $\pm$ 0.992 & 9.300 $\pm$ 1.308 & 9.302 $\pm$ 1.588 & 6.351 $\pm$ 1.376 & 2.206 $\pm$ 0.264 & 5.176 $\pm$ 0.818 \\
AdaLead & 10.510 $\pm$ 1.409 & 6.866 $\pm$ 1.883 & 4.886 $\pm$ 1.221 & \underline{24.584 $\pm$ 5.407} & \underline{32.806 $\pm$ 3.168} & 9.557 $\pm$ 1.650 & 2.642 $\pm$ 0.115 & 8.132 $\pm$ 3.066 \\
GFN-AL-$\delta$CS  & 9.831 $\pm$ 0.325 & 5.414 $\pm$ 0.843 & 6.364 $\pm$ 0.616 & 23.630 $\pm$ 4.945 & 24.415 $\pm$ 3.535 & \textbf{11.372 $\pm$ 2.705} & 2.948 $\pm$ 0.048 & 7.685 $\pm$ 1.071 \\
MLDE & 2.896 $\pm$ 0.479 & 4.056 $\pm$ 0.710 & 1.280 $\pm$ 0.916 & 3.168 $\pm$ 0.420 & 4.218 $\pm$ 1.267 & 1.924 $\pm$ 0.638 & 1.080 $\pm$ 0.104 & 2.444 $\pm$ 0.699 \\
ProSpero & \textbf{13.118 $\pm$ 1.647} & \textbf{16.666 $\pm$ 3.931} & \textbf{9.236 $\pm$ 1.931} & \textbf{42.328 $\pm$ 5.149} & \textbf{64.146 $\pm$ 4.232} & 10.416 $\pm$ 2.879 & \underline{3.224 $\pm$ 0.872} & \textbf{16.256 $\pm$ 3.193} \\
\midrule
SILO & \underline{12.320 $\pm$ 1.411} & \underline{15.892 $\pm$ 1.610} & \underline{8.516 $\pm$ 1.395} & 20.040 $\pm$ 0.825 & 32.432 $\pm$ 20.070 & \underline{10.879 $\pm$ 2.013} & \textbf{6.803 $\pm$ 1.872} & \underline{14.815 $\pm$ 2.996} \\
\bottomrule
\end{tabular}
}
\end{table}

\begin{table}[htbp]
\centering
\caption{Average diversity between top-100 generated sequences generated by all methods on 8 benchmark datasets across 10 active learning rounds. Reported values are the mean $\pm$ standard deviation over 5 independent runs. \textbf{Bold}: the best per each task. \underline{Underline}: second-best.}
\label{tab:div_rerun}
\footnotesize
\setlength{\tabcolsep}{3pt}
\renewcommand{\arraystretch}{0.9}
\resizebox{\linewidth}{!}{
\begin{tabular}{lcccccccc}
\toprule
Method & AAV & AMIE & E4B & GFP & LGK & Pab1 & TEM & UBE2I \\
\midrule
PEX & \underline{7.069 $\pm$ 1.250} & 6.996 $\pm$ 0.905 & 5.164 $\pm$ 1.022 & 10.045 $\pm$ 1.117 & 8.042 $\pm$ 2.237 & \underline{5.253 $\pm$ 0.659} & 4.016 $\pm$ 0.439 & 7.002 $\pm$ 0.629 \\
AdaLead & 6.766 $\pm$ 1.026 & 7.921 $\pm$ 3.262 & \underline{6.651 $\pm$ 0.749} & \underline{27.435 $\pm$ 9.213} & \underline{19.442 $\pm$ 3.454} & 3.736 $\pm$ 1.473 & 3.853 $\pm$ 0.130 & \textbf{11.461 $\pm$ 3.550} \\
GFN-AL-$\delta$CS  & \textbf{10.743 $\pm$ 0.965} & \underline{9.287 $\pm$ 1.706} & \textbf{7.368 $\pm$ 1.735} & \textbf{36.140 $\pm$ 6.136} & \textbf{43.118 $\pm$ 6.043} & \textbf{6.262 $\pm$ 1.382} & \underline{4.214 $\pm$ 0.096} & \underline{11.015 $\pm$ 0.513} \\
MLDE & 2.395 $\pm$ 0.381 & 4.950 $\pm$ 0.580 & 1.757 $\pm$ 1.055 & 4.805 $\pm$ 0.690 & 6.885 $\pm$ 1.781 & 2.331 $\pm$ 0.688 & 1.893 $\pm$ 0.205 & 4.064 $\pm$ 0.999 \\
ProSpero & 5.565 $\pm$ 0.317 & \textbf{12.404 $\pm$ 0.658} & 3.815 $\pm$ 0.184 & 10.298 $\pm$ 1.520 & 12.519 $\pm$ 1.205 & 3.683 $\pm$ 0.369 & \textbf{4.629 $\pm$ 0.828} & 9.294 $\pm$ 1.592 \\
\midrule
SILO & 2.624 $\pm$ 0.362 & 3.288 $\pm$ 0.484 & 2.619 $\pm$ 0.339 & 3.427 $\pm$ 1.140 & 6.497 $\pm$ 6.769 & 2.784 $\pm$ 0.219 & 2.523 $\pm$ 0.491 & 3.496 $\pm$ 0.537 \\
\bottomrule
\end{tabular}
}
\end{table}

\begin{table}[H]
\centering
\caption{Median fitness of top-50 generated sequences generated all methods on 8 benchmark datasets across 10 active learning rounds. Reported values are the mean $\pm$ standard deviation over 5 independent runs. \textbf{Bold}: the best per each task. \underline{Underline}: second-best.}
\label{tab:median_rerun}
\footnotesize
\setlength{\tabcolsep}{3pt}
\renewcommand{\arraystretch}{0.9}
\resizebox{\linewidth}{!}{
\begin{tabular}{lllllllll}
\toprule
Method & AAV & AMIE & E4B & GFP & LGK & Pab1 & TEM & UBE2I \\
\midrule
PEX           & 0.598$\pm$0.010 & \underline{0.233$\pm$0.001 } & 7.869$\pm$0.051 & 3.596$\pm$0.001 & 0.032$\pm$0.001 & 1.473$\pm$0.226 & 1.228$\pm$0.000 & \underline{2.986$\pm$0.002} \\
AdaLead       & \underline{0.650$\pm$0.013} & 0.227$\pm$0.010  & 7.863$\pm$0.062 & 3.570$\pm$0.002 & 0.033$\pm$0.003 & 1.445$\pm$0.159 & 1.205$\pm$0.002 & 2.974$\pm$0.002 \\

GFN-AL-$\delta$CS & 0.634$\pm$0.007 & 0.181$\pm$0.013  & 7.911$\pm$0.052 & 3.575$\pm$0.002 & 0.019$\pm$0.002 & 1.480$\pm$0.325 & 1.203$\pm$0.005 & 2.981$\pm$0.002 \\

MLDE          & 0.531$\pm$0.005 & 0.229$\pm$0.002  & 7.745$\pm$0.006 & 3.578$\pm$0.001 & 0.022$\pm$0.001 & 0.894$\pm$0.030 & \underline{1.227$\pm$0.000}& 2.981$\pm$0.000 \\

ProSpero      & 0.627$\pm$0.025 & 0.232$\pm$0.005  & \underline{8.056$\pm$0.011} & \underline{3.611$\pm$0.003 }& \underline{0.041$\pm$0.001} & \underline{1.503$\pm$0.425} & 1.208$\pm$0.012 & 2.985$\pm$0.002 \\
\midrule
SILO  & \textbf{0.726$\pm$0.014} & \textbf{0.259$\pm$0.005}  & \textbf{8.146$\pm$0.058}& \textbf{3.617$\pm$0.001} & \textbf{0.044$\pm$0.002} & \textbf{1.854$\pm$0.356} & \textbf{1.232$\pm$0.001} & \textbf{2.996$\pm$0.002} \\
\bottomrule
\end{tabular}
}
\end{table}

\subsection {Statistical significance tests}
We conducted two-sided paired \textit{t}-test over the five aligned seeds against the strongest reproduced baseline, selected separately for each task and metric. We additionally report paired mean differences, 95\% confidence intervals, Hedges’ $g_z$, and Holm-adjusted \textit{p}-values to account for multiple hypothesis testing (Tables \ref{tab:max_paired_statistics}) and  \ref{tab:mean_paired_statistics}). 
 \begin{table}[th]
    \centering
    \caption{Paired comparison of maximum fitness between SILO and the strongest reproduced baseline for each benchmark using five aligned seeds. We report the paired mean difference ($\Delta$), two-sided 95\% confidence interval (CI), Hedges' $g_z$, the raw two-sided paired $t$-test $p$-value, and the Holm-adjusted $p$-value across the eight benchmarks. An asterisk indicates raw or Holm-adjusted $p<0.05$.}
    \footnotesize
    \setlength{\tabcolsep}{4pt}
    \begin{tabular}{l l c c c c c}
    \toprule
    \textbf{Protein} &
    \textbf{Baseline} &
    \textbf{$\Delta$} &
    \textbf{95\% CI} &
    \textbf{Hedges' $g_z$} &
    \textbf{Raw $p$} &
    \textbf{Holm $p$} \\
    \midrule
    AAV   & AdaLead           & 0.049 & [0.023, 0.076] & 1.897 & 0.006* & 0.042* \\
    
    AMIE  & PEX               & 0.017 & [0.014, 0.020] & 5.221 & 0.0001* & $8 \times10^{-4}$*  \\
    
    E4B   & ProSpero          & 0.027 & [-0.046, 0.100] & 0.368 & 0.361 & 1.000 \\
    
    GFP   & ProSpero          & 0.004 & [-0.002, 0.010] & 0.643 & 0.146 & 0.730\\
    
    LGK   & ProSpero          & 0.002 & [-0.001, 0.006] & 0.559 & 0.192 & 0.768\\
    
    Pab1  & PEX               & 0.094 & [-0.621, 0.811] & 0.131 & 0.731 & 1.000 \\
    
    TEM   & PEX               &  $2.72\times10^{-5}$  & [-0.003, 0.003] & 0.008 & 0.982 & 1.000 \\
    
    UBE2I & GFN-AL-$\delta$CS & 0.006 & [0.002, 0.010] & 1.468 & 0.014* & 0.084 \\
    \bottomrule
    \end{tabular}
    \label{tab:max_paired_statistics}
    \end{table}

   \begin{table}[th]
    \centering
    \caption{Paired statistical comparison of top-100 mean fitness between SILO and the strongest reproduced baseline for each benchmark using five aligned random seeds. We report the paired mean difference $\Delta$, two-sided 95\% confidence interval (CI), Hedges' $g_z$, the raw two-sided paired $t$-test $p$-value, and the Holm-adjusted $p$-value across the eight benchmarks. An asterisk indicates raw or Holm-adjusted $p<0.05$.}
    \footnotesize
    \setlength{\tabcolsep}{4pt}
    \begin{tabular}{l l c c c c c}
    \toprule
    \textbf{Protein} &
    \textbf{Baseline} &
    \textbf{$\Delta$} &
    \textbf{95\% CI} &
    \textbf{Hedges' $g_z$} &
    \textbf{Raw $p$} &
    \textbf{Holm $p$}\\
    \midrule
    AAV   & AdaLead   & 0.066  & [0.041, 0.090] & 2.691 & 0.001* & 0.007*  \\
    
    AMIE  & PEX               & 0.024 & [0.018, 0.031] & 3.810 & $4.40\times10^{-4*}$ & 0.004*  \\
    
    E4B   & ProSpero          & 0.087 & [0.019, 0.154] & 1.287 & 0.022* & 0.084  \\
    
    GFP   & ProSpero          & 0.006 & [0.001, 0.011] & 1.318 & 0.021* & 0.084 \\
    
    LGK   & ProSpero          & 0.003 & [-$4.83\times10^{-4}$, 0.007] & 0.865 & 0.072 & 0.144  \\
    
    Pab1  & ProSpero               & 0.344 & [-0.255, 0.943] & 0.570 & 0.186 & 0.186 \\
    
    TEM   & PEX               & 0.004 & [0.002, 0.005] & 3.106 & $9.68\times10^{-4*}$ & 0.007* \\
    
    UBE2I & PEX & 0.009 & [0.005, 0.013] & 2.334 & 0.002* & 0.010*  \\
    \bottomrule
    \end{tabular}
    \label{tab:mean_paired_statistics}
    \end{table}
    
\subsection {Early-round protein design}
\begin{table}[H]
\centering
\caption{Maximum fitness achieved by all methods on 8 benchmark datasets for the first 5 active learning rounds. Reported values are the mean $\pm$ standard deviation over 5 independent runs. \textbf{Bold}: the best per each task. \underline{Underline}: second-best}
\label{tab:max_fit_early}
\scriptsize
\setlength{\tabcolsep}{3pt}
\renewcommand{\arraystretch}{0.9}
\resizebox{\linewidth}{!}{
\begin{tabular}{lllllllll}
\toprule
Method & AAV & AMIE & E4B & GFP & LGK & Pab1 & TEM & UBE2I \\
\midrule

PEX           & 0.610$\pm$0.013& \underline{0.237 $\pm$ 0.002}& 7.999$\pm$ 0.074& 3.597 $\pm$ 0.001& 0.032 $\pm$ 0.002& 1.254 $\pm$ 0.151& \underline{1.231 $\pm$ 0.001}& \underline{2.989 $\pm$ 0.002}\\

AdaLead       & \underline{0.634$\pm$ 0.013}& 0.230 $\pm$ 0.005& 8.000$\pm$ 0.063& 3.583 $\pm$0.002& 0.034 $\pm$ 0.004& 1.240 $\pm$ 0.107& 1.228$\pm$0.000& 2.985 $\pm$ 0.003\\

GFN-AL-$\delta$CS & 0.605$\pm$0.007& 0.227$\pm$0.004& 8.019$\pm$0.032& 3.590$\pm$0.001& 0.030$\pm$0.003& 1.330$\pm$0.263& 1.228$\pm$0.000& 2.987$\pm$0.000\\

MLDE          & 0.543$\pm$0.008& 0.235$\pm$0.002& 7.857$\pm$0.036& 3.585$\pm$0.003& 0.024$\pm$0.001& 0.935$\pm$0.000& 1.228$\pm$0.000& 2.985$\pm$0.000\\

ProSpero & 
0.617 $\pm$ 0.025& 0.234 $\pm$ 0.005& \underline{8.021 $\pm$ 0.048}& \underline{3.604 $\pm$ 0.004}& \textbf{0.037 $\pm$ 0.002}& \underline{1.411$\pm$ 0.356}& 1.229$\pm$ 0.001& 2.987 $\pm$ 0.002\\

\midrule

SILO & {\textbf{0.640 $\pm$ 0.020}} & {\textbf{0.251 $\pm$ 0.005}}& {\textbf{8.114 $\pm$ 0.058}}& {\textbf{3.608 $\pm$ 0.002}}& {\underline{0.038$\pm$ 0.003}}& {\textbf{1.698 $\pm$ 0.402}}& {\textbf{1.232 $\pm$ 0.001}} & {\textbf{2.993 $\pm$ 0.001}}\\

\bottomrule
\end{tabular}
}
\end{table}

\begin{table}[h]
\centering
\caption{Mean fitness of top-100 sequences generated by all methods on 8 benchmark datasets for first 5 active learning rounds. Reported values are the mean $\pm$ standard deviation over 5 independent runs. \textbf{Bold}: the best per each task. \underline{Underline}: second-best}
\label{tab:mean_fit_early}
\scriptsize
\setlength{\tabcolsep}{3pt}
\renewcommand{\arraystretch}{0.9}
\resizebox{\linewidth}{!}{
\begin{tabular}{lllllllll}
\toprule
Method & AAV & AMIE & E4B & GFP & LGK & Pab1 & TEM & UBE2I \\
\midrule
PEX           & 0.558$\pm$0.010& \underline{0.228$\pm$ 0.002}& 7.738$\pm$ 0.053& 3.588 $\pm$ 0.001& 0.027 $\pm$ 0.002& 1.022 $\pm$ 0.008& 1.220 $\pm$ 0.001 & \underline{2.983$\pm$ 0.001}\\

AdaLead       & \underline{0.574$\pm$ 0.012}& 0.218 $\pm$ 0.006 & 7.772 $\pm$ 0.063& 3.563 $\pm$0.003& 0.029 $\pm$ 0.005& 1.115 $\pm$ 0.115 & 1.204$\pm$0.000& 2.967$\pm$ 0.003\\

GFN-AL-$\delta$CS & 0.564$\pm$0.003& 0.176$\pm$0.012& 7.785$\pm$0.013& 3.568$\pm$0.002& 0.017$\pm$0.003& 1.077$\pm$0.106& 1.204$\pm$0.002& 2.977$\pm$0.002\\

MLDE          & 0.518$\pm$0.008& 0.227$\pm$0.001& 7.744$\pm$0.004& 3.576$\pm$0.001& 0.021$\pm$0.000& 0.860$\pm$0.004& \underline{1.226$\pm$0.001}& 2.980$\pm$0.000\\

ProSpero & 
0.566 $\pm$ 0.019& 0.225$\pm$ 0.005& \underline{7.841 $\pm$ 0.045}& \underline{3.599 $\pm$ 0.004}& \underline{0.034 $\pm$ 0.002}& \underline{1.239 $\pm$ 0.246}& 1.087 $\pm$ 0.038 & 2.978 $\pm$ 0.004\\

\midrule
SILO & {\textbf{0.606 $\pm$ 0.011}}& {\textbf{0.244 $\pm$ 0.005}}& {\textbf{8.044 $\pm$ 0.087}}& {\textbf{3.605 $\pm$ 0.003}}& {\textbf{0.037 $\pm$ 0.003}}& {\textbf{1.504 $\pm$ 0.384}}& {\textbf{1.231 $\pm$ 0.001}}& {\textbf{2.991 $\pm$ 0.001}}\\
\bottomrule
\end{tabular}
}
\end{table}

\newpage
\subsection {Low-labeled data regime}
\begin{table}[H]
\centering
\caption{
Maximum fitness and mean fitness of top-100 generated sequences by all methods under the low-data setting using 10\% of the
initial $D_0$ dataset. Results are reported as mean $\pm$ standard deviation over 5 independent runs.
\textbf{Bold}: best result for each task and metric.
\underline{Underline}: second-best result.
}
\label{tab:max_mean_fitness_low_data_0.1}
\scriptsize
\setlength{\tabcolsep}{3pt}
\renewcommand{\arraystretch}{0.95}
\resizebox{\linewidth}{!}{
\begin{tabular}{lcccccccc}
\toprule
& \multicolumn{4}{c}{Maximum fitness}
& \multicolumn{4}{c}{Mean fitness} \\
\cmidrule(lr){2-5}
\cmidrule(lr){6-9}
Method
& UBE2I & TEM & LGK & Pab1
& UBE2I & TEM & LGK & Pab1 \\
\midrule

PEX
& 2.990 $\pm$ 0.004
& \underline{1.228 $\pm$ 0.005}
& \underline{0.035 $\pm$ 0.002}
& 1.536 $\pm$ 0.340
& 2.984 $\pm$ 0.003
& 1.203 $\pm$ 0.012
& 0.032 $\pm$ 0.001
& 1.330 $\pm$ 0.304
\\

AdaLead
& 2.986 $\pm$ 0.004
& 1.220 $\pm$ 0.009
& 0.034 $\pm$ 0.001
& 1.239 $\pm$ 0.301
& 2.976 $\pm$ 0.006
& 1.196 $\pm$ 0.003
& 0.031 $\pm$ 0.002
& 1.136 $\pm$ 0.246
\\

GFN-AL-$\delta$CS
& 2.986 $\pm$ 0.004
& 1.219 $\pm$ 0.009
& 0.017 $\pm$ 0.0019
& 0.611 $\pm$ 0.172
& 2.975 $\pm$ 0.009
& 1.200 $\pm$ 0.013
& -0.024 $\pm$ 0.000
& 0.161 $\pm$ 0.007
\\

MLDE
& 2.986 $\pm$ 0.001
& \underline{1.228 $\pm$ 0.000}
& 0.024 $\pm$ 0.000
& 0.937 $\pm$ 0.003
& 2.982 $\pm$ 0.000
& \underline{1.222 $\pm$ 0.005}
& 0.022 $\pm$ 0.000
& 0.928 $\pm$ 0.004
\\

ProSpero
& \underline{2.992 $\pm$ 0.002}
& \textbf{1.229 $\pm$ 0.001}
& \textbf{0.042 $\pm$ 0.001}
& \textbf{1.739 $\pm$ 0.312}
& \underline{2.986 $\pm$ 0.001}
& 1.161 $\pm$ 0.046
& \underline{0.040 $\pm$ 0.001}
& \textbf{1.611 $\pm$ 0.272}
\\

\midrule

SILO
& \textbf{2.993 $\pm$ 0.005}
& \textbf{1.229 $\pm$ 0.001}
& \textbf{0.042 $\pm$ 0.001}
& \underline{1.591 $\pm$ 0.474}
& \textbf{2.992 $\pm$ 0.005}
& \textbf{1.228 $\pm$ 0.000}
& \textbf{0.041 $\pm$ 0.001}
& \underline{1.450 $\pm$ 0.455}
\\

\bottomrule
\end{tabular}
}
\end{table}

\begin{table}[H]
\centering
\caption{
Average novelty and diversity of the top-100 generated sequences
under the low-data setting using 10\% of the initial $D_0$
dataset. Results are reported as mean $\pm$ standard deviation over 5 independent runs.
\textbf{Bold}: best result per task and metric.
\underline{Underline}: second-best result.
}
\label{tab:low_data_novelty_diversity_0.1}
\setlength{\tabcolsep}{3pt}
\footnotesize
\renewcommand{\arraystretch}{0.95}
\resizebox{\linewidth}{!}{%
\begin{tabular}{lcccccccc}
\toprule
& \multicolumn{4}{c}{Novelty}
& \multicolumn{4}{c}{Diversity} \\
\cmidrule(lr){2-5}
\cmidrule(lr){6-9}
Method
& UBE2I & TEM & LGK & Pab1
& UBE2I & TEM & LGK & Pab1 \\
\midrule

PEX
& 5.440$\pm$1.798
& 1.160$\pm$0.272
& 10.406$\pm$2.155
& 6.715$\pm$1.940
& \underline{7.807$\pm$1.644}
& 3.313$\pm$0.296
& 7.894$\pm$1.603
& \underline{5.970$\pm$1.227} \\

AdaLead
& 4.660$\pm$2.292
& 0.910$\pm$0.082
& 52.266$\pm$10.176
& 11.418$\pm$1.657
& 7.498$\pm$2.207
& 3.638$\pm$0.136
& \underline{26.002$\pm$3.422}
& 4.826$\pm$1.896 \\

GFN-AL-$\delta$CS
& 5.376$\pm$1.782
& 1.564$\pm$0.749
& \textbf{425.002$\pm$0.406}
& \textbf{71.226$\pm$0.251}
& 5.812$\pm$1.382
& \underline{4.045$\pm$0.672}
& \textbf{424.433$\pm$0.080}
& \textbf{71.123$\pm$0.024} \\

MLDE
& 5.192$\pm$1.800
& \textbf{3.979$\pm$0.493}
& 4.892$\pm$0.826
& 2.056$\pm$0.286
& 5.165$\pm$0.718
& 1.872$\pm$0.179
& 7.520$\pm$1.427
& 1.825$\pm$0.650 \\

ProSpero
& \textbf{17.584$\pm$3.784}
& \underline{2.608$\pm$0.407}
& \underline{60.100$\pm$6.931}
& \underline{11.440$\pm$0.829}
& \textbf{10.371$\pm$0.842}
& \textbf{4.133$\pm$0.411}
& \underline{16.152$\pm$3.274}
& 3.626$\pm$0.150 \\

\midrule

SILO
& \underline{10.580$\pm$3.255}
& 1.456$\pm$0.564
& 21.844$\pm$3.006
& 10.680$\pm$1.346
& 2.587$\pm$0.580
& 1.704$\pm$0.120
& 4.081$\pm$0.701
& 2.984$\pm$0.135 \\

\bottomrule
\end{tabular}%
}
\end{table}

\begin{table}[H]
\centering
\caption{
Maximum fitness and mean fitness under the low-data setting using 20\% of the
initial $D_0$ dataset. Results are reported over 5 independent runs.
\textbf{Bold}: best result for each task and metric.
\underline{Underline}: second-best result.
}
\label{tab:max_mean_fitness_low_data_0.2}
\scriptsize
\setlength{\tabcolsep}{3pt}
\renewcommand{\arraystretch}{0.95}
\resizebox{\linewidth}{!}{
\begin{tabular}{lcccccccc}
\toprule
& \multicolumn{4}{c}{Maximum fitness}
& \multicolumn{4}{c}{Mean fitness} \\
\cmidrule(lr){2-5}
\cmidrule(lr){6-9}
Method
& UBE2I & TEM & LGK & Pab1
& UBE2I & TEM & LGK & Pab1 \\
\midrule

PEX
& \textbf{2.992 $\pm$ 0.001}
& \underline{1.229 $\pm$ 0.003}
& 0.034 $\pm$ 0.001
& \underline{1.690 $\pm$ 0.294}
& \underline{2.987 $\pm$ 0.000}
& 1.215 $\pm$ 0.008
& 0.031 $\pm$ 0.001
& 1.449 $\pm$ 0.366
\\

AdaLead
& 2.989 $\pm$ 0.002
& 1.227 $\pm$ 0.001
& 0.036 $\pm$ 0.002
& \textbf{1.719 $\pm$ 0.465}
& 2.979 $\pm$ 0.002
& 1.201 $\pm$ 0.000
& 0.033 $\pm$ 0.003
& \textbf{1.625 $\pm$ 0.428}
\\

GFN-AL-$\delta$CS
& \underline{2.990 $\pm$ 0.003}
& 1.224 $\pm$ 0.007
& 0.018 $\pm$ 0.001
& 0.735 $\pm$ 0.098
& 2.981 $\pm$ 0.008
& 1.213 $\pm$ 0.008
& -0.016 $\pm$ 0.000
& 0.185 $\pm$ 0.025
\\

MLDE
& 2.985 $\pm$ 0.000
& 1.228 $\pm$ 0.000
& 0.024 $\pm$ 0.001
& 0.939 $\pm$ 0.002
& 2.981 $\pm$ 0.000
& \underline{1.224 $\pm$ 0.003}
& 0.022 $\pm$ 0.000
& 0.930 $\pm$ 0.001
\\

ProSpero
& \textbf{2.992 $\pm$ 0.004}
& \textbf{1.230 $\pm$ 0.001}
& \underline{0.042 $\pm$ 0.002}
& 1.420 $\pm$ 0.316
& 2.985 $\pm$ 0.003
& 1.152 $\pm$ 0.036
& \underline{0.041 $\pm$ 0.003}
& 1.336 $\pm$ 0.261
\\

\midrule

SILO
& \textbf{2.992 $\pm$ 0.005}
& \underline{1.229 $\pm$ 0.001}
& \textbf{0.043 $\pm$ 0.001}
& 1.674 $\pm$ 0.424
& \textbf{2.992 $\pm$ 0.004}
& \textbf{1.228 $\pm$ 0.001}
& \textbf{0.042 $\pm$ 0.001}
& \underline{1.600 $\pm$ 0.424}
\\

\bottomrule
\end{tabular}
}
\end{table}

\begin{table}[H]
\centering
\caption{
Average novelty and diversity of the top-100 generated sequences
under the low-data setting using 20\% of the initial $D_0$
dataset across five independent runs.
Values represent mean $\pm$ standard deviation.
\textbf{Bold}: best result per task and metric.
\underline{Underline}: second-best result.
}
\label{tab:low_data_20_novelty_diversity}
\setlength{\tabcolsep}{3pt}
\footnotesize
\renewcommand{\arraystretch}{0.95}
\resizebox{\linewidth}{!}{%
\begin{tabular}{lcccccccc}
\toprule
& \multicolumn{4}{c}{Novelty}
& \multicolumn{4}{c}{Diversity} \\
\cmidrule(lr){2-5}
\cmidrule(lr){6-9}
Method
& UBE2I & TEM & LGK & Pab1
& UBE2I & TEM & LGK & Pab1 \\
\midrule

PEX
& 4.873$\pm$0.631
& 1.507$\pm$0.698
& 8.546$\pm$0.913
& 7.514$\pm$0.459
& 7.905$\pm$1.175
& 3.464$\pm$0.411
& 7.838$\pm$0.965
& \underline{5.865$\pm$0.675} \\

AdaLead
& 3.596$\pm$0.493
& 0.621$\pm$0.131
& 45.868$\pm$10.567
& 10.158$\pm$0.880
& 7.092$\pm$1.089
& 3.102$\pm$0.235
& \underline{24.923$\pm$7.244}
& 4.406$\pm$1.106 \\

GFN-AL-$\delta$CS
& 6.846$\pm$1.777
& 2.507$\pm$0.667
& \textbf{424.669$\pm$0.596}
& \textbf{71.160$\pm$0.112}
& \underline{8.908$\pm$2.035}
& \underline{4.755$\pm$0.583}
& \textbf{424.510$\pm$0.087}
& \textbf{71.129$\pm$0.036} \\

MLDE
& 5.320$\pm$1.071
& \textbf{4.562$\pm$0.439}
& 4.792$\pm$0.742
& 1.982$\pm$0.137
& 5.529$\pm$0.691
& 2.227$\pm$0.233
& 6.272$\pm$1.397
& 1.702$\pm$0.200 \\

ProSpero
& \textbf{16.210$\pm$4.653}
& \underline{3.358$\pm$0.922}
& \underline{59.084$\pm$9.199}
& 11.604$\pm$1.622
& \textbf{11.095$\pm$1.276}
& \textbf{5.006$\pm$1.087}
& 13.506$\pm$1.878
& 3.578$\pm$0.186 \\

\midrule

SILO
& \underline{9.864$\pm$2.990}
& 3.239$\pm$2.407
& 20.696$\pm$3.424
& \underline{12.132$\pm$2.062}
& 2.796$\pm$0.479
& 2.006$\pm$0.487
& 4.434$\pm$0.283
& 3.227$\pm$0.307 \\

\bottomrule
\end{tabular}%
}
\end{table}

\begin{table}[H]
\centering
\caption{
Maximum fitness and mean fitness under the low-data setting using 50\% of the
initial $D_0$ dataset. Results are reported over 5 independent runs.
\textbf{Bold}: best result for each task and metric.
\underline{Underline}: second-best result.
}
\label{tab:max_mean_fitness_low_data_0.5}
\scriptsize
\setlength{\tabcolsep}{3pt}
\renewcommand{\arraystretch}{0.95}
\resizebox{\linewidth}{!}{
\begin{tabular}{lcccccccc}
\toprule
& \multicolumn{4}{c}{Maximum fitness}
& \multicolumn{4}{c}{Mean fitness} \\
\cmidrule(lr){2-5}
\cmidrule(lr){6-9}
Method
& UBE2I & TEM & LGK & Pab1
& UBE2I & TEM & LGK & Pab1 \\
\midrule

PEX
& \underline{2.993 $\pm$ 0.002}
& \underline{1.229 $\pm$ 0.001}
& 0.035 $\pm$ 0.001
& \underline{1.740 $\pm$ 0.437}
& \underline{2.987 $\pm$ 0.000}
& 1.220 $\pm$ 0.004
& 0.032 $\pm$ 0.002
& 1.510 $\pm$ 0.305
\\

AdaLead
& 2.987 $\pm$ 0.003
& 1.228 $\pm$ 0.001
& 0.038 $\pm$ 0.002
& \textbf{1.909 $\pm$ 0.190}
& 2.975 $\pm$ 0.003
& 1.204 $\pm$ 0.001
& 0.035 $\pm$ 0.002
& \textbf{1.726 $\pm$ 0.223}
\\

GFN-AL-$\delta$CS
& 2.990 $\pm$ 0.002
& 1.228 $\pm$ 0.000
& 0.019 $\pm$ 0.000
& 0.808 $\pm$ 0.077
& 2.981 $\pm$ 0.005
& 1.206 $\pm$ 0.003
& -0.003 $\pm$ 0.001
& 0.239 $\pm$ 0.026
\\

MLDE
& 2.986 $\pm$ 0.000
& \underline{1.229 $\pm$ 0.000}
& 0.025 $\pm$ 0.000
& 0.946 $\pm$ 0.011
& 2.982 $\pm$ 0.000
& \underline{1.223 $\pm$ 0.004}
& 0.023 $\pm$ 0.001
& 0.923 $\pm$ 0.005
\\

ProSpero
& 2.992 $\pm$ 0.003
& \underline{1.229 $\pm$ 0.000}
& \underline{0.042 $\pm$ 0.002}
& 1.550 $\pm$ 0.317
& 2.986 $\pm$ 0.002
& 1.101 $\pm$ 0.042
& \underline{0.041 $\pm$ 0.002}
& 1.453 $\pm$ 0.316
\\

\midrule

SILO
& \textbf{2.995 $\pm$ 0.006}
& \textbf{1.230 $\pm$ 0.002}
& \textbf{0.044 $\pm$ 0.001}
& 1.683 $\pm$ 0.279
& \textbf{2.994 $\pm$ 0.005}
& \textbf{1.229 $\pm$ 0.001}
& \textbf{0.043 $\pm$ 0.001}
& \underline{1.572 $\pm$ 0.293}
\\

\bottomrule
\end{tabular}
}
\end{table}

\begin{table}[H]
\centering
\caption{
Average novelty and diversity of the top-100 generated sequences
under the low-data setting using 50\% of the initial $D_0$
dataset across five independent runs.
Values represent mean $\pm$ standard deviation.
\textbf{Bold}: best result per task and metric.
\underline{Underline}: second-best result.
}
\label{tab:low_data_50_novelty_diversity}
\setlength{\tabcolsep}{3pt}
\footnotesize
\renewcommand{\arraystretch}{0.95}
\resizebox{\linewidth}{!}{%
\begin{tabular}{lcccccccc}
\toprule
& \multicolumn{4}{c}{Novelty}
& \multicolumn{4}{c}{Diversity} \\
\cmidrule(lr){2-5}
\cmidrule(lr){6-9}
Method
& UBE2I & TEM & LGK & Pab1
& UBE2I & TEM & LGK & Pab1 \\
\midrule

PEX
& 6.157$\pm$0.879
& 1.525$\pm$0.233
& 9.768$\pm$0.977
& 7.312$\pm$3.337
& 8.350$\pm$0.374
& 3.515$\pm$0.262
& 7.903$\pm$1.375
& \underline{5.133$\pm$0.629} \\

AdaLead
& 4.309$\pm$0.717
& 0.578$\pm$0.075
& 38.750$\pm$11.402
& 10.318$\pm$4.507
& \underline{8.692$\pm$1.009}
& 2.993$\pm$0.123
& \underline{20.172$\pm$7.390}
& 4.785$\pm$2.180 \\

GFN-AL-$\delta$CS
& 6.006$\pm$1.454
& 1.223$\pm$0.800
& \textbf{424.602$\pm$0.194}
& \textbf{71.174$\pm$0.274}
& 8.164$\pm$2.749
& \underline{3.898$\pm$1.040}
& \textbf{424.528$\pm$0.056}
& \textbf{71.128$\pm$0.0210} \\

MLDE
& 5.056$\pm$0.013
& \textbf{5.050$\pm$0.933}
& 4.904$\pm$0.645
& 2.528$\pm$0.198
& 5.461$\pm$0.839
& 2.275$\pm$0.296
& 6.658$\pm$1.621
& 2.705$\pm$0.579 \\

ProSpero
& \textbf{18.938$\pm$4.909}
& \underline{2.864$\pm$0.818}
& \underline{64.126$\pm$5.210}
& \underline{11.188$\pm$1.895}
& \textbf{11.420$\pm$0.936}
& \textbf{4.380$\pm$0.550}
& 12.877$\pm$2.704
& 3.911$\pm$0.239 \\

\midrule

SILO
& \underline{11.908$\pm$3.113}
& 2.752$\pm$2.999
& 22.291$\pm$2.457
& 10.358$\pm$1.146
& 2.871$\pm$0.616
& 1.848$\pm$0.431
& 3.799$\pm$0.527
& 3.086$\pm$0.280 \\

\bottomrule
\end{tabular}%
}
\end{table}

\subsection {Noisy proxy setting}

\begin{table}[H] 
\centering 
\caption{Results under noisy proxy setting. Maximum fitness and mean fitness of top-100 sequences by all methods on four benchmark datasets across 10 active learning rounds under the $-5$ signal-to-noise setting. Reported values represent mean $\pm$ standard deviation across 5 independent runs. \textbf{Bold}: best result for each task and metric. \underline{Underline}: second-best result. } \label{tab:max_mean_fitness_noisy_5} \scriptsize \setlength{\tabcolsep}{3pt} \renewcommand{\arraystretch}{0.95} \resizebox{\linewidth}{!}{ \begin{tabular}{lcccccccc} \toprule & \multicolumn{4}{c}{Maximum fitness} & \multicolumn{4}{c}{Mean fitness} \\ \cmidrule(lr){2-5} \cmidrule(lr){6-9} Method & AMIE & E4B & GFP & AAV & AMIE & E4B & GFP & AAV \\ \midrule PEX & \underline{0.242 $\pm$ 0.001} & 8.002 $\pm$ 0.074 & 3.603 $\pm$ 0.002 & 0.626 $\pm$ 0.014 & \underline{0.234 $\pm$ 0.001} & 7.826 $\pm$ 0.037 & 3.595 $\pm$ 0.002 & 0.573 $\pm$ 0.015 \\ AdaLead & 0.235 $\pm$ 0.001 & 8.002 $\pm$ 0.040 & 3.586 $\pm$ 0.001 & 0.688 $\pm$ 0.027 & 0.227 $\pm$ 0.001 & 7.805 $\pm$ 0.034 & 3.570 $\pm$ 0.001 & 0.641 $\pm$ 0.027 \\ GFN-AL-$\delta$CS & 0.229 $\pm$ 0.002 & 7.952 $\pm$ 0.064 & 3.590 $\pm$ 0.002 & 0.567 $\pm$ 0.003 & 0.212 $\pm$ 0.003 & 7.700 $\pm$ 0.031 & 3.580 $\pm$ 0.001 & 0.516 $\pm$ 0.005 \\ MLDE & 0.233 $\pm$ 0.002 & 7.836 $\pm$ 0.063 & 3.582 $\pm$ 0.001 & 0.525 $\pm$ 0.000 & 0.227 $\pm$ 0.001 & 7.738 $\pm$ 0.021 & 3.576 $\pm$ 0.000 & 0.506 $\pm$ 0.003 \\ ProSpero & \underline{0.242 $\pm$ 0.003} & \textbf{8.139 $\pm$ 0.019} & \underline{3.616 $\pm$ 0.003} & \underline{0.695 $\pm$ 0.015} & 0.233 $\pm$ 0.003 & \underline{8.074 $\pm$ 0.023} & \underline{3.611 $\pm$ 0.003} & \underline{0.655 $\pm$ 0.018} \\ 
\midrule SILO & \textbf{0.260 $\pm$ 0.002} & \underline{8.118 $\pm$ 0.040} & \textbf{3.620 $\pm$ 0.002} & \textbf{0.704 $\pm$ 0.035} & \textbf{0.257 $\pm$ 0.003} & \textbf{8.088 $\pm$ 0.044} & \textbf{3.618 $\pm$ 0.002} & \textbf{0.673 $\pm$ 0.026} \\ \bottomrule \end{tabular} } 
\end{table}

\begin{table}[H]
\centering
\caption{
Average novelty and diversity of the top-100 sequences
by all methods on four benchmark datasets across 10 active learning rounds under the $-5$ signal-to-noise setting. Reported values represent mean $\pm$ standard deviation across 5 independent runs.
\textbf{Bold}: best result per task and metric.
\underline{Underline}: second-best result.
}
\label{tab:noisy_novelty_diversity_5}
\setlength{\tabcolsep}{3pt}
\footnotesize
\renewcommand{\arraystretch}{0.95}
\resizebox{\linewidth}{!}{%
\begin{tabular}{lcccccccc}
\toprule
& \multicolumn{4}{c}{Novelty}
& \multicolumn{4}{c}{Diversity} \\
\cmidrule(lr){2-5}
\cmidrule(lr){6-9}
Method
& AMIE & E4B & GFP & AAV
& AMIE & E4B & GFP & AAV \\
\midrule

PEX
& 3.934$\pm$0.242
& 2.932$\pm$0.561
& 7.914$\pm$1.998
& 5.414$\pm$0.974
& 6.442$\pm$0.738
& 5.161$\pm$0.823
& 11.277$\pm$2.310
& \underline{7.024$\pm$0.726} \\

AdaLead
& 4.135$\pm$0.421
& 3.175$\pm$0.509
& \underline{22.606$\pm$2.789}
& \underline{10.558$\pm$1.527}
& \underline{8.264$\pm$0.700}
& \underline{5.742$\pm$0.579}
& \textbf{34.636$\pm$2.832}
& 6.137$\pm$1.434 \\

GFN-AL-$\delta$CS
& 1.985$\pm$0.210
& 3.796$\pm$0.509
& 17.610$\pm$5.068
& 4.696$\pm$0.384
& 4.835$\pm$0.341
& \textbf{6.512$\pm$0.539}
& \underline{24.559$\pm$5.178}
& \textbf{8.072$\pm$0.401} \\

MLDE
& 2.985$\pm$0.074
& 1.690$\pm$0.139
& 2.268$\pm$0.197
& 1.682$\pm$0.116
& 3.343$\pm$0.429
& 2.586$\pm$0.260
& 4.003$\pm$0.264
& 2.673$\pm$0.160 \\

ProSpero
& \textbf{17.916$\pm$1.868}
& \textbf{9.970$\pm$1.982}
& \textbf{48.762$\pm$3.168}
& \textbf{14.472$\pm$0.740}
& \textbf{12.485$\pm$0.732}
& 4.329$\pm$0.510
& 15.842$\pm$0.420
& 6.229$\pm$0.748 \\

\midrule

SILO
& \underline{14.172$\pm$2.557}
& \underline{8.532$\pm$1.336}
& \underline{18.253$\pm$1.623}
& \underline{9.936$\pm$1.744}
& 2.917$\pm$0.355
& 2.709$\pm$0.217
& 3.836$\pm$0.813
& 3.164$\pm$0.206 \\

\bottomrule
\end{tabular}%
}
\end{table}

\begin{table}[H] \centering \caption{Results under noisy proxy setting. Maximum fitness and mean fitness of top-100 sequences by all methods on four benchmark datasets across 10 active learning rounds under the $-15$ signal-to-noise setting. Reported values represent mean $\pm$ standard deviation across 5 independent runs. \textbf{Bold}: best result for each task and metric. \underline{Underline}: second-best result.} \label{tab:max_mean_fitness_noisy_15} \scriptsize \setlength{\tabcolsep}{3pt} \renewcommand{\arraystretch}{0.95} \resizebox{\linewidth}{!}{ \begin{tabular}{lcccccccc} \toprule & \multicolumn{4}{c}{Maximum fitness} & \multicolumn{4}{c}{Mean fitness} \\ \cmidrule(lr){2-5} \cmidrule(lr){6-9} Method & AMIE & E4B & GFP & AAV & AMIE & E4B & GFP & AAV \\ \midrule PEX & \underline{0.242 $\pm$ 0.001} & 7.999 $\pm$ 0.033 & 3.602 $\pm$ 0.002 & 0.586 $\pm$ 0.007 & \underline{0.233 $\pm$ 0.001} & 7.783 $\pm$ 0.026 & 3.595 $\pm$ 0.002 & 0.531 $\pm$ 0.006 \\ AdaLead & 0.237 $\pm$ 0.007 & 7.941 $\pm$ 0.081 & 3.583 $\pm$ 0.002 & 0.574 $\pm$ 0.041 & 0.228 $\pm$ 0.003 & 7.717 $\pm$ 0.045 & 3.562 $\pm$ 0.002 & 0.529 $\pm$ 0.037 \\ GFN-AL-$\delta$CS & 0.226 $\pm$ 0.001 & 7.866 $\pm$ 0.036 & 3.588 $\pm$ 0.004 & 0.532 $\pm$ 0.006 & 0.205 $\pm$ 0.002 & 7.277 $\pm$ 0.128 & 3.576 $\pm$ 0.001 & 0.482 $\pm$ 0.003 \\ MLDE & 0.233 $\pm$ 0.002 & 7.840 $\pm$ 0.050 & 3.582 $\pm$ 0.001 & 0.531 $\pm$ 0.010 & 0.227 $\pm$ 0.001 & 7.728 $\pm$ 0.020 & 3.575 $\pm$ 0.000 & 0.495 $\pm$ 0.006 \\ ProSpero & 0.238 $\pm$ 0.004 & \underline{8.100 $\pm$ 0.022} & \underline{3.616 $\pm$ 0.003} & \underline{0.649 $\pm$ 0.031} & 0.226 $\pm$ 0.004 & \underline{7.953 $\pm$ 0.076} & \underline{3.609 $\pm$ 0.002} & \underline{0.598 $\pm$ 0.030} 
\\ \midrule SILO & \textbf{0.259 $\pm$ 0.004} & \textbf{8.119 $\pm$ 0.089} & \textbf{3.617 $\pm$ 0.002} & \textbf{0.689 $\pm$ 0.024} & \textbf{0.256 $\pm$ 0.004} & \textbf{8.090 $\pm$ 0.093} & \textbf{3.615 $\pm$ 0.002} & \textbf{0.644 $\pm$ 0.011} \\ \bottomrule \end{tabular} } \end{table}

\begin{table}[H]
\centering
\caption{
Average novelty and diversity of the top-100 sequences
by all methods on four benchmark datasets across 10 active learning rounds under the $-15$ signal-to-noise setting. Reported values represent mean $\pm$ standard deviation across 5 independent runs.
\textbf{Bold}: best result per task and metric.
\underline{Underline}: second-best result.
}
\label{tab:noisy_15_novelty_diversity}
\setlength{\tabcolsep}{3pt}
\footnotesize
\renewcommand{\arraystretch}{0.95}
\resizebox{\linewidth}{!}{%
\begin{tabular}{lcccccccc}
\toprule
& \multicolumn{4}{c}{Novelty}
& \multicolumn{4}{c}{Diversity} \\
\cmidrule(lr){2-5}
\cmidrule(lr){6-9}
Method
& AMIE & E4B & GFP & AAV
& AMIE & E4B & GFP & AAV \\
\midrule

PEX
& 3.621$\pm$0.416
& 2.582$\pm$0.214
& 7.620$\pm$1.146
& 3.412$\pm$0.292
& 5.961$\pm$0.735
& 4.699$\pm$0.315
& 11.368$\pm$1.185
& 5.483$\pm$0.443 \\

AdaLead
& 4.662$\pm$1.262
& 3.243$\pm$0.518
& \underline{26.125$\pm$3.414}
& 4.948$\pm$1.917
& \underline{7.881$\pm$0.464}
& \underline{5.930$\pm$0.778}
& \textbf{42.165$\pm$5.277}
& 5.849$\pm$0.922 \\

GFN-AL-$\delta$CS
& 1.843$\pm$0.268
& 3.543$\pm$0.262
& 16.782$\pm$5.940
& 3.520$\pm$0.150
& 4.803$\pm$0.403
& \textbf{5.960$\pm$0.153}
& \underline{24.311$\pm$7.52}
& \textbf{6.645$\pm$0.283} \\

MLDE
& 2.879$\pm$0.133
& 1.644$\pm$0.260
& 1.908$\pm$0.197
& 1.319$\pm$0.060
& 3.089$\pm$0.259
& 2.538$\pm$0.504
& 3.472$\pm$0.344
& 2.086$\pm$0.471 \\

ProSpero
& \textbf{16.274$\pm$3.813}
& \underline{7.028$\pm$1.661}
& \textbf{49.168$\pm$4.558}
& \textbf{13.546$\pm$1.954}
& \textbf{14.019$\pm$1.878}
& 5.081$\pm$0.223
& 17.787$\pm$2.054
& \underline{6.276$\pm$0.425} \\

\midrule

SILO
& \underline{13.880$\pm$1.225}
& \textbf{8.492$\pm$2.085}
& 17.966$\pm$2.793
& \underline{8.950$\pm$1.576}
& 3.145$\pm$0.954
& 2.441$\pm$0.286
& 3.700$\pm$0.379
& 3.150$\pm$0.170 \\

\bottomrule
\end{tabular}%
}
\end{table}

\begin{table}[H] \centering \caption{Results under noisy proxy setting. Maximum fitness and mean fitness of top-100 sequences by all methods on four benchmark datasets across 10 active learning rounds under the $-25$ signal-to-noise setting. Reported values represent mean $\pm$ standard deviation across 5 independent runs. \textbf{Bold}: best result for each task and metric. \underline{Underline}: second-best result.} \label{tab:max_mean_fitness_noisy_25} \scriptsize \setlength{\tabcolsep}{3pt} \renewcommand{\arraystretch}{0.95} \resizebox{\linewidth}{!}{ \begin{tabular}{lcccccccc} \toprule & \multicolumn{4}{c}{Maximum fitness} & \multicolumn{4}{c}{Mean fitness} \\ \cmidrule(lr){2-5} \cmidrule(lr){6-9} Method & AMIE & E4B & GFP & AAV & AMIE & E4B & GFP & AAV \\ \midrule PEX & \underline{0.241 $\pm$ 0.001} & 7.956 $\pm$ 0.036 & 3.604 $\pm$ 0.002 & \underline{0.564 $\pm$ 0.005} & \underline{0.232 $\pm$ 0.002} & 7.744 $\pm$ 0.018 & 3.597 $\pm$ 0.002 & \underline{0.514 $\pm$ 0.005} \\ AdaLead & 0.237 $\pm$ 0.004 & 7.929 $\pm$ 0.074 & 3.583 $\pm$ 0.002 & 0.512 $\pm$ 0.004 & 0.225 $\pm$ 0.001 & 7.602 $\pm$ 0.092 & 3.557 $\pm$ 0.003 & 0.438 $\pm$ 0.011 \\ GFN-AL-$\delta$CS & 0.227 $\pm$ 0.003 & 7.773 $\pm$ 0.053 & 3.583 $\pm$ 0.004 & 0.525 $\pm$ 0.011 & 0.201 $\pm$ 0.001 & 5.674 $\pm$ 0.127 & 3.566 $\pm$ 0.003 & 0.455 $\pm$ 0.005 \\ MLDE & 0.234 $\pm$ 0.003 & 7.885 $\pm$ 0.081 & 3.582 $\pm$ 0.000 & 0.529 $\pm$ 0.009 & 0.226 $\pm$ 0.002 & 7.725 $\pm$ 0.023 & 3.575 $\pm$ 0.000 & 0.490 $\pm$ 0.008 \\ ProSpero & 0.233 $\pm$ 0.003 & \underline{7.997 $\pm$ 0.106} & \underline{3.613 $\pm$ 0.003} & 0.559 $\pm$ 0.032 & 0.217 $\pm$ 0.003 & \underline{7.785 $\pm$ 0.141} & \underline{3.607 $\pm$ 0.002} & 0.493 $\pm$ 0.022 
\\ \midrule SILO & \textbf{0.257 $\pm$ 0.001} & \textbf{8.086 $\pm$ 0.030} & \textbf{3.618 $\pm$ 0.001} & \textbf{0.670 $\pm$ 0.035} & \textbf{0.254 $\pm$ 0.000} & \textbf{8.056 $\pm$ 0.034} & \textbf{3.616 $\pm$ 0.001} & \textbf{0.635 $\pm$ 0.033} \\ \bottomrule \end{tabular} } \end{table}

\begin{table}[H]
\centering
\caption{
Average novelty and diversity of the top-100 sequences
by all methods on four benchmark datasets across 10 active learning rounds under the $-25$ signal-to-noise setting. Reported values represent mean $\pm$ standard deviation across 5 independent runs.
\textbf{Bold}: best result per task and metric.
\underline{Underline}: second-best result.
}
\label{tab:noisy_25_novelty_diversity}
\setlength{\tabcolsep}{3pt}
\footnotesize
\renewcommand{\arraystretch}{0.95}
\resizebox{\linewidth}{!}{%
\begin{tabular}{lcccccccc}
\toprule
& \multicolumn{4}{c}{Novelty}
& \multicolumn{4}{c}{Diversity} \\
\cmidrule(lr){2-5}
\cmidrule(lr){6-9}
Method
& AMIE & E4B & GFP & AAV
& AMIE & E4B & GFP & AAV \\
\midrule

PEX
& 3.262$\pm$0.646
& 2.126$\pm$0.246
& 8.456$\pm$1.077
& 2.338$\pm$0.248
& 5.605$\pm$0.701
& 3.912$\pm$0.362
& 11.754$\pm$1.755
& 4.268$\pm$0.509 \\

AdaLead
& 4.482$\pm$0.300
& 3.623$\pm$0.608
& \underline{29.833$\pm$4.268}
& 3.392$\pm$0.061
& \underline{8.638$\pm$0.448}
& \underline{6.261$\pm$0.807}
& \textbf{48.738$\pm$6.171}
& 6.418$\pm$0.092 \\

GFN-AL-$\delta$CS
& 1.553$\pm$0.145
& 1.684$\pm$0.188
& 21.454$\pm$5.738
& 3.572$\pm$0.250
& 4.339$\pm$0.175
& \textbf{7.308$\pm$0.339}
& \underline{33.207$\pm$6.993}
&\underline{ 6.817$\pm$0.468} \\

MLDE
& 3.076$\pm$0.153
& 1.666$\pm$0.200
& 1.932$\pm$0.168
& 1.180$\pm$0.136
& 3.153$\pm$0.299
& 2.515$\pm$0.625
& 3.458$\pm$0.305
& 1.955$\pm$0.430 \\

ProSpero
& \underline{12.078$\pm$3.181}
& \underline{6.324$\pm$1.752}
& \textbf{53.108$\pm$1.373}
& \underline{7.372$\pm$1.412}
& \textbf{14.366$\pm$0.581}
& 5.752$\pm$0.193
& 19.187$\pm$2.451
& \textbf{7.507$\pm$0.452} \\

\midrule

SILO
& \textbf{14.315$\pm$1.927}
& \textbf{7.828$\pm$1.053}
& 19.990$\pm$2.111
& \textbf{9.238$\pm$1.602}
& 3.465$\pm$0.373
& 2.632$\pm$0.236
& 3.870$\pm$0.936
& 3.196$\pm$0.233 \\

\bottomrule
\end{tabular}%
}
\end{table}

\subsection {Ablation studies}

\begin{table}[H]
\caption{Ablation results across all 8 protein fitness landscapes,
reporting the maximum fitness achieved over 10
active-learning rounds. Reported values represent the mean $\pm$  standard
deviation across 5 independent runs. The ablations examine
the contributions of sequence generation strategy, search and candidate
selection strategies, and iterative policy learning. Values in brackets denote wildtype $(x_{WT})$ fitness. \textbf{Bold}: the best per each task.}
\label{tab:ablation_complete}

\centering
\small
\setlength{\tabcolsep}{2pt}
\renewcommand{\arraystretch}{1.1}

\begin{adjustbox}{max width=\linewidth}
\begin{tabular}{@{}l*{8}{c}@{}}
\toprule
Landscape
& AAV (0.500) & AMIE (0.224) & E4B (7.743) & GFP (3.572) & LGK (0.020) & Pab1 (0.84) & TEM (1.229) & UBE2I (2.978) \\

\midrule
\multicolumn{9}{c}{\textbf{Sequence generation strategy}} \\
\midrule

Random baseline 

& 0.530$\pm$0.015 & 0.229 $\pm$ 0.002 & 6.970 $\pm$1.156 & 3.600$\pm$0.010 & 0.027 $\pm$0.001 & 1.018$\pm$ 0.103 & 1.229 $\pm$ 0.000 & 2.984 $\pm$ 0.000 \\

\midrule
\multicolumn{9}{c}{\textbf{Sampling and selection}} \\
\midrule

SIL + BS + w/o AFS score
& 0.549$\pm$0.026 & 0.230$\pm$0.006 & 7.642$\pm$0.289  & 3.593$\pm$0.006 &  0.026$\pm$0.002  & 0.942$\pm$0.090
& 1.228$\pm$0.000 & 2.986$\pm$ 0.001\\

SIL + BS + with AFS score
& 0.587$\pm$0.028 & 0.232 $\pm$ 0.007 & 7.706 $\pm$0.173 & 3.602$\pm$0.012 & 0.026 $\pm$0.001 & 1.022$\pm$ 0.146 & 1.228 $\pm$ 0.000 & 2.987 $\pm$ 0.001 \\

SIL + SBS + w/o AFS score
& 0.694$\pm$0.019 & 0.258 $\pm$0.003 & 8.140 $\pm$0.062 & 3.615$\pm$0.001 & 0.042$\pm$0.001
& 1.558 $\pm$ 0.134 & 1.232 $\pm$ 0.000 & \textbf{2.999 $\pm$ 0.001} \\

\midrule
\multicolumn{9}{c}{\textbf{Impact of Learning}} \\
\midrule

Frozen policy +  BS + AFS score
& 0.551$\pm$0.024 & 0.227$\pm$0.005  & 7.604$\pm$0.269  & 3.595$\pm$0.010 & 0.027$\pm$0.003 & 0.901$\pm$0.045 & 1.228$\pm$0.000 & 2.987$\pm$0.001 \\

Frozen policy + SBS + AFS score
& 0.706$\pm$0.013 & 0.256 $\pm$ 0.003 & 8.149 $\pm$ 0.041 & 3.615$\pm$0.001 & 0.042 $\pm$0.001 & 1.530$\pm$0.268 & 1.233 $\pm$0.002 & 2.997 $\pm$ 0.001 \\

\midrule
SILO (SBS + AFS score + SIL)
& \textbf{0.747 $\pm$ 0.014} & \textbf{0.262 $\pm$ 0.004} &\textbf{ 8.169 $\pm$ 0.052}
& \textbf{3.619 $\pm$ 0.002} & \textbf{0.045 $\pm$ 0.003} & \textbf{1.894 $\pm$ 0.348} & \textbf{1.233 $\pm$ 0.003} & 2.997 $\pm$ 0.003 \\

\bottomrule
\end{tabular}
\end{adjustbox}
\end{table}

\begin{table}[H]
\caption{Ablation results across all 8 protein fitness landscapes,
reporting the mean fitness of top-100 sequences achieved over 10
active-learning rounds. Reported values represent the mean $\pm$  standard
deviation across 5 independent runs. The ablations examine
the contributions of sequence generation strategy, search and candidate
selection strategies, and iterative policy learning. Values in brackets denote wildtype $(x_{WT})$ fitness. \textbf{Bold}: the best per each task.}
\label{tab:mean_ablation_complete}

\centering
\small
\setlength{\tabcolsep}{2pt}
\renewcommand{\arraystretch}{1.1}

\begin{adjustbox}{max width=\linewidth}
\begin{tabular}{@{}l*{8}{c}@{}}
\toprule
Landscape
& AAV (0.500) & AMIE (0.224) & E4B (7.743) & GFP (3.572) & LGK (0.020) & Pab1 (0.84) & TEM (1.229) & UBE2I (2.978) \\

\midrule
\multicolumn{9}{c}{\textbf{Sequence generation strategy}} \\
\midrule

Random baseline 

& 0.526$\pm$0.014 & 0.225 $\pm$ 0.001 & 6.722 $\pm$1.138 & 3.599$\pm$0.010 & 0.024 $\pm$0.001 & 0.997$\pm$ 0.010 & 1.228 $\pm$ 0.000 & 2.982 $\pm$ 0.000 \\

\midrule
\multicolumn{9}{c}{\textbf{Sampling and selection}} \\
\midrule

SIL + BS + w/o AFS score
& 0.548$\pm$0.025 & 0.230$\pm$0.006 & 7.601$\pm$0.319  & 3.592$\pm$0.006 &  0.025$\pm$0.000  & 0.929$\pm$0.093
& 1.217$\pm$0.024 & 2.985$\pm$ 0.001\\

SIL + BS + with AFS score
& 0.582$\pm$0.028 & 0.231 $\pm$ 0.007 & 7.641 $\pm$0.248 & 3.600$\pm$0.011 & 0.025 $\pm$0.001 & 1.002$\pm$ 0.143 & 1.222 $\pm$ 0.011 & 2.986 $\pm$ 0.001 \\

SIL + SBS + w/o AFS score
& 0.673$\pm$0.014 & 0.253 $\pm$0.004 & 8.107 $\pm$0.067 & 3.614$\pm$0.000 & 0.041$\pm$0.001
& 1.467 $\pm$ 0.111 & 1.231 $\pm$ 0.000 & \textbf{2.998 $\pm$ 0.001} \\

\midrule
\multicolumn{9}{c}{\textbf{Impact of Learning}} \\
\midrule

Frozen policy +  BS + AFS score
& 0.549$\pm$0.022 & 0.227$\pm$0.004  & 7.589$\pm$0.271  & 3.594$\pm$0.010 & 0.027$\pm$0.003 & 0.898$\pm$0.042 & 1.215$\pm$0.020 & 2.986$\pm$0.001 \\

Frozen policy + SBS + AFS score
& 0.669$\pm$0.024 & 0.251 $\pm$ 0.003 & 8.106 $\pm$ 0.060 & 3.613$\pm$0.001 & 0.041 $\pm$0.001 & 1.383$\pm$0.207 & 1.231 $\pm$0.001 & 2.995 $\pm$ 0.001 \\

\midrule
SILO (SBS + AFS score + SIL)
& \textbf{0.727 $\pm$ 0.014} & \textbf{0.259 $\pm$ 0.005} &\textbf{ 8.148 $\pm$ 0.058}
& \textbf{3.617 $\pm$ 0.001} & \textbf{0.044 $\pm$ 0.002} & \textbf{1.856 $\pm$ 0.355} & \textbf{1.232 $\pm$ 0.001} & 2.996 $\pm$ 0.002 \\

\bottomrule
\end{tabular}
\end{adjustbox}
\end{table}

\begin{table}[H]
\caption{Ablation results across all 8 protein fitness landscapes,
reporting the novelty achieved over 10
active-learning rounds. Reported values represent the mean $\pm$  standard
deviation across 5 independent runs. The ablations examine
the contributions of sequence generation strategy, search and candidate selection strategies, and iterative policy learning. Values in brackets denote wildtype $(x_{WT})$ fitness. \textbf{Bold}: the best per each task.}
\label{tab:novelty_ablation_complete}

\centering
\small
\setlength{\tabcolsep}{2pt}
\renewcommand{\arraystretch}{1.1}

\begin{adjustbox}{max width=\linewidth}
\begin{tabular}{@{}l*{8}{c}@{}}
\toprule
Landscape
& AAV (0.500) & AMIE (0.224) & E4B (7.743) & GFP (3.572) & LGK (0.020) & Pab1 (0.84) & TEM (1.229) & UBE2I (2.978) \\

\midrule
\multicolumn{9}{c}{\textbf{{Sequence generation strategy}}} \\
\midrule

Random baseline 

& 3.268$\pm$1.436 & 1.866 $\pm$ 0.043 & 1.230 $\pm$0.252 & 13.052$\pm$6.317 & 6.452 $\pm$ 1.919 & 4.928$\pm$ 2.663 & 1.478 $\pm$ 0.521 & 2.700 $\pm$ 0.071 \\

\midrule
\multicolumn{9}{c}{\textbf{Sampling and selection}} \\
\midrule

SIL + BS + w/o AFS score
& 3.524$\pm$1.648 & 4.802$\pm$4.159 & 1.473$\pm$1.049  & 9.064$\pm$2.930 &  6.994$\pm$3.277  & 2.916$\pm$2.847
& 1.354$\pm$0.229 & 4.950$\pm$ 1.067\\

SIL + BS + with AFS score
& 5.556$\pm$1.327 & 5.216 $\pm$ 4.248 & 1.723 $\pm$0.945 & 10.099$\pm$3.812 & 6.614 $\pm$3.116 & 4.360$\pm$ 3.408 & 1.372 $\pm$ 0.248 & 6.078 $\pm$ 1.461 \\

SIL + SBS + w/o AFS score
& 9.584$\pm$0.685 & 14.406 $\pm$2.543 & 8.354 $\pm$1.870 & 19.360$\pm$2.542 & 22.386$\pm$2.356
& 10.512$\pm$ 1.742 & \textbf{6.830 $\pm$ 1.916} & 13.434 $\pm$ 1.524 \\

\midrule
\multicolumn{9}{c}{\textbf{Impact of Learning}} \\
\midrule

Frozen policy +  BS + AFS score
& 3.736$\pm$1.398 & 3.636$\pm$2.210  & 1.030$\pm$0.375  & 12.356$\pm$4.095 & 8.408$\pm$2.931 & 1.564$\pm$0.507 & 1.230$\pm$0.213 & 5.808$\pm$1.465 \\

Frozen policy + SBS + AFS score
& 10.388 $\pm$0.973 & 14.218 $\pm$ 2.179 & 7.916 $\pm$ 1.114 & 18.612 $\pm$ 3.984 & 19.360 $\pm$2.179 & 10.384$\pm$ 1.911 & 6.102 $\pm$1.515 & 13.366 $\pm$ 2.197 \\

\midrule
SILO (SBS + AFS score + SIL)
& \textbf{12.320 $\pm$ 1.411} & \textbf{15.892 $\pm$ 1.610} &\textbf{ 8.516 $\pm$ 1.395}
& \textbf{20.040 $\pm$ 0.825} & \textbf{32.432 $\pm$ 20.070} & \textbf{10.798 $\pm$ 2.013} & 6.803 $\pm$ 1.872 & \textbf{14.815 $\pm$ 2.992} \\

\bottomrule
\end{tabular}
\end{adjustbox}
\end{table}

\begin{table}[H]
\caption{Ablation results across all 8 protein fitness landscapes,
reporting the diversity achieved over 10 active-learning rounds. Values represent the mean $\pm$  standard deviation across 5 independent runs. The ablations examine
the contributions of sequence generation strategy, search and candidate selection strategies, and iterative policy learning. Values in brackets denote wildtype $(x_{WT})$ fitness. \textbf{Bold}: the best per each task.}
\label{tab:diversity_ablation_complete}

\centering
\small
\setlength{\tabcolsep}{2pt}
\renewcommand{\arraystretch}{1.1}

\begin{adjustbox}{max width=\linewidth}
\begin{tabular}{@{}l*{8}{c}@{}}
\toprule
Landscape
& AAV (0.500) & AMIE (0.224) & E4B (7.743) & GFP (3.572) & LGK (0.020) & Pab1 (0.84) & TEM (1.229) & UBE2I (2.978) \\

\midrule
\multicolumn{9}{c}{\textbf{{Sequence generation strategy}}} \\
\midrule

Random baseline 

& 1.518$\pm$1.147 & 3.292 $\pm$ 0.669 & 2.227 $\pm$0.500 & 2.139$\pm$ 0.382 & \textbf{7.740 $\pm$ 2.957} & 1.531$\pm$ 0.888 & 2.055$\pm$ 0.074 & \textbf{4.237 $\pm$ 0.927} \\

\midrule
\multicolumn{9}{c}{\textbf{Sampling and selection}} \\
\midrule

SIL + BS + w/o AFS score
& 0.665$\pm$ 0.419 & 1.432$\pm$ 0.779 & 1.044$\pm$0.434  & 1.745$\pm$ 0.708 &  2.250$\pm$0.895  & 0.600$\pm$0.543
& 0.798$\pm$0.172 & 1.593$\pm$ 0.420\\

SIL + BS + with AFS score
& 1.701$\pm$1.156 & 1.614 $\pm$ 0.938 & 1.083 $\pm$0.334 & 1.912$\pm$ 0.821 & 2.139 $\pm$0.786 & 1.076$\pm$ 0.844 & 0.825 $\pm$ 0.220 & 1.983 $\pm$ 0.323 \\

SIL + SBS + w/o AFS score
& 2.477$\pm$0.167 & \textbf{3.642$\pm$ 0.885} & \textbf{2.927 $\pm$ 0.194} & 2.861$\pm$ 0.191 & 3.716$\pm$ 0.588
& 2.665$\pm$ 0.029 & 3.081 $\pm$ 0.266 & 3.197 $\pm$ 0.560 \\

\midrule
\multicolumn{9}{c}{\textbf{Impact of Learning}} \\
\midrule

Frozen policy +  BS + AFS score
& 0.656$\pm$ 0.425 & 1.115$\pm$ 0.651  & 0.728$\pm$0.254  & 2.159$\pm$ 0.435 & 2.335$\pm$0.581 & 0.081$\pm$0.094 & 0.564 $\pm$0.089 & 1.594$\pm$0.507 \\

Frozen policy + SBS + AFS score
& 2.616 $\pm$ 0.225 & 3.474 $\pm$ 0.291 & 2.875 $\pm$ 0.351 & 2.778 $\pm$ 0.220 & 3.896 $\pm$ 0.494 & 2.669$\pm$ 0.386 & \textbf{3.123 $\pm$ 0.409} & 3.013 $\pm$ 0.292 \\

\midrule
SILO (SBS + AFS score + SIL)
& \textbf{2.624 $\pm$ 0.362} & 3.288 $\pm$ 0.484 & 2.619 $\pm$ 0.339
& \textbf{3.427 $\pm$ 1.140} & 6.497 $\pm$ 6.769 & \textbf{2.784 $\pm$ 0.219} & 2.523 $\pm$ 0.491 & 3.496 $\pm$ 0.537 \\

\bottomrule
\end{tabular}
\end{adjustbox}
\end{table}

\begin{table}[h]
\centering
\caption{Paired statistical comparison of max-fitness values attained by full SILO and frozen-policy SBS for each benchmark using the five shared seeds. We report the paired mean difference ($\Delta$), 95\% confidence interval (CI), Hedges' $g_z$, the raw paired $t$-test $p$-value, and the Holm-adjusted $p$-value. An asterisk indicates raw or Holm-adjusted $p<0.05$.}
\footnotesize
\setlength{\tabcolsep}{4pt}
\begin{tabular}{l c c c c c }
\toprule
\textbf{Protein} &
\textbf{$\Delta$} &
\textbf{95\% CI} &
\textbf{Hedges' $g_z$} &
\textbf{Raw $p$} &
\textbf{Holm $p$} \\
\midrule
AAV              & 0.040 & [0.023, 0.057] & 2.365 & 0.002* & 0.016* \\
AMIE                 & 0.006 & [0.001, 0.010] & 1.321 & 0.020* & 0.114 \\
E4B             & 0.020 & [-0.066, 0.108] & 0.233 & 0.548 & 1.000  \\
GFP             & 0.004 & [0.001, 0.006] & 1.721 & 0.008* & 0.056  \\
LGK             & 0.002 & [-0.001, 0.006] & 0.549 & 0.199 & 0.796  \\
Pab1                 & 0.363 & [0.096, 0.630] & 1.352 & 0.019* & 0.114  \\
TEM                 & -0.000 & [-0.002, 0.002] & -0.045 & 0.905 & 1.000  \\
UBE2I  & 0.000 & [-0.002, 0.004] & 0.193 & 0.618 & 1.000 \\
\bottomrule
\end{tabular}
\label{tab:frozen_sbs_paired_statistics}
\end{table}
\noindent
\begin{table}[h]
\centering
\caption{Paired statistical comparison of max fitness values attained by SILO against the SIL-beam search-AFS variant for each benchmark using the five shared random seeds. We report the paired mean difference ($\Delta$), 95\% confidence interval (CI), Hedges' $g_z$, the raw paired $t$-test $p$-value, and the Holm-adjusted $p$-value. An asterisk indicates raw or Holm-adjusted $p<0.05$.}
\footnotesize
\setlength{\tabcolsep}{4pt}
\begin{tabular}{l c c c c c}
\toprule
\textbf{Protein} &
\textbf{$\Delta$} &
\textbf{95\% CI} &
\textbf{Hedges' $g_z$} &
\textbf{Raw $p$} &
\textbf{Holm $p$} \\
\midrule
AAV             & 0.159 & [0.133, 0.185] & 6.066 & $7.11 \times10^{-5*}$ & $5.6 \times 10^{-4*}$ \\
AMIE               & 0.030 & [0.016, 0.045] & 2.146 & 0.003*  & 0.012* \\
E4B           & 0.463 & [0.271, 0.654] & 2.403 & 0.002* & 0.010* \\
GFP           & 0.017 & [0.001, 0.032] & 1.115 & 0.035* & 0.035* \\
LGK           & 0.019 & [0.012, 0.025] & 3.111 & $9.6 \times 10^{-4*}$ & 0.007* \\
Pab1               & 0.871 & [0.465, 1.270] & 2.134 & 0.003* & 0.012* \\
TEM                & 0.004 & [0.002, 0.006] & 2.204 & 0.003* & 0.012* \\
UBE2I & 0.010 & [0.007, 0.013] & 3.105 & $9.6 \times 10^{-4*}$ & 0.007*  \\
\bottomrule
\end{tabular}
\label{tab:beam_search_paired_statistics}
\end{table}

\subsection {Hyperparameter sensitivity analysis}
\paragraph{Influence of varying number of top trajectories $P$ for policy training}
For the number of top oracle evaluated trajectories used for training policy, $P = 50$ achieves the highest mean maximum fitness on six landscapes, ties at the displayed precision on TEM, and is within 0.001 of the best setting on UBE2I (Table \ref{tab:varying_P}). Relative to $P = 25$, it improves the mean on six landscapes and ties on two. Increasing $P$ further to 75 does not provide a consistent additional benefit, suggesting that $P = 50$ provides a reasonable balance between the number and quality of imitation trajectories.

\paragraph{Influence of varying $\beta$ beam size within stochastic beam search} Increasing the beam width from $\beta = 16$ to $\beta = 32$ improves the reported mean on seven landscapes and ties on UBE2I (Table \ref{tab:varying_beta}). Increasing it further $\beta =$64 to produces only small, task-dependent changes without a consistent improvement across the suite. We therefore retain  as a practical balance between optimization performance and sampling cost.

\paragraph{Influence of varying $\gamma_1$ and $\gamma_2$ UCB coefficients} Performance is generally stable across the evaluated uncertainty coefficients. The default $(\gamma_1,\gamma_2) = (0.1,1.0)$ is best or tied at the displayed precision on seven of eight landscapes and is within $0.002$ of the best result on UBE2I (Table \ref{tab:varying_gamma}). These results show that the selected setting is locally stable within the tested range.

\begin{table}[H]
\centering
\caption{Effect of number of top oracle evaluated trajectories ($P$) for policy training on SILO's performance on 8 benchmark datasets across 10 active learning rounds (maximum fitness). Reported values are the mean $\pm$  standard deviation over 5 independent runs. Values in brackets denote wildtype $(x_{WT})$ fitness. \textbf{Bold}: the best per each task. \underline{Underline}: second-best}
\label{tab:varying_P}
\scriptsize
\setlength{\tabcolsep}{3pt}
\renewcommand{\arraystretch}{0.9}
\resizebox{\linewidth}{!}{
\begin{tabular}{lllllllll}
\toprule
Parameter & AAV (0.500) & AMIE (0.224) & E4B (7.743) & GFP (3.572) & LGK (0.020) & Pab1 (0.84) & TEM (1.229) & UBE2I (2.978) \\
\midrule

$P=$25           & \underline{0.725 $\pm$ 0.027} & \underline{0.260 $\pm$ 0.006} &\underline{ 8.119 $\pm$ 0.047} & \underline{3.618 $\pm$ 0.002} & 0.042 $\pm$ 0.001 & 1.481 $\pm$ 0.436 & 1.233 $\pm$ 0.001 & \underline{2.997 $\pm$ 0.002} \\

$P=$75           & 0.709 $\pm$ 0.031 & 0.258 $\pm$ 0.004 & 8.107 $\pm$ 0.043 & 3.616 $\pm$ 0.001 & \underline{0.043 $\pm$ 0.001} & \underline{1.665 $\pm$ 0.337} & 1.233 $\pm$ 0.002 & \textbf{2.998 $\pm$ 0.002} \\

\midrule

$P=$50 (SILO)   & \textbf{0.747 $\pm$ 0.014} & \textbf{0.262 $\pm$ 0.004} & \textbf{8.169 $\pm$ 0.052} & \textbf{3.619 $\pm$ 0.002} & \textbf{0.045 $\pm$ 0.003} & \textbf{1.894 $\pm$ 0.348} & 1.233 $\pm$ 0.003 & \underline{2.997 $\pm$ 0.003} \\
\bottomrule
\end{tabular}
}
\end{table}

\begin{table}[H]
\centering
\caption{Effect of varying length of beam search ($\beta$) within stochastic beam search on SILO's performance on 8 benchmark datasets across 10 active learning rounds (maximum fitness). Reported values are the mean $\pm$  standard deviation over 5 independent runs. Values in brackets denote wildtype $(x_{WT})$ fitness. \textbf{Bold}: the best per each task. \underline{Underline}: second-best}
\label{tab:varying_beta}
\scriptsize
\setlength{\tabcolsep}{3pt}
\renewcommand{\arraystretch}{0.9}
\resizebox{\linewidth}{!}{
\begin{tabular}{lllllllll}
\toprule
Parameter & AAV (0.500) & AMIE (0.224) & E4B (7.743) & GFP (3.572) & LGK (0.020) & Pab1 (0.84) & TEM (1.229) & UBE2I (2.978) \\
\midrule

$\beta$=16           & 0.676 $\pm$ 0.027 & \underline{0.258$\pm$0.004} & 8.143 $\pm$ 0.028 & 3.617 $\pm$ 0.002 & 0.042 $\pm$ 0.002 & 1.508 $\pm$ 0.283 & 1.231 $\pm$ 0.001 & \underline{2.997 $\pm$ 0.003} \\

$\beta$=64           & \underline{0.728 $\pm$ 0.021} & \textbf{0.262 $\pm$ 0.005} & \textbf{8.171 $\pm$ 0.027} & \underline{3.618 $\pm$ 0.002} & \underline{0.044 $\pm$ 0.002} & \textbf{1.898 $\pm$ 0.180} & \textbf{1.234 $\pm$ 0.001} & \textbf{2.999 $\pm$ 0.002} \\

\midrule

$\beta$=32 (SILO)   & \textbf{0.747 $\pm$ 0.014} & \textbf{0.262 $\pm$ 0.004} & \underline{8.169 $\pm$ 0.052 }& \textbf{3.619 $\pm$ 0.002} & \textbf{0.045 $\pm$ 0.003} & \underline{1.894 $\pm$ 0.348} & \underline{1.233 $\pm$ 0.003} & \underline{2.997 $\pm$ 0.003} \\
\bottomrule
\end{tabular}
}
\end{table}

\begin{table}[H]
\centering
\caption{Sensitivity analysis over all four combinations of the UCB uncertainty coefficients $\gamma_1$ and $\gamma_2$ on SILO's performance on 8 benchmark datasets across 10 active learning rounds (maximum fitness). Reported values are the mean $\pm$  standard deviation over 5 independent runs. Values in brackets denote wildtype $(x_{WT})$ fitness. \textbf{Bold}: the best per each task. \underline{Underline}: second-best}
\label{tab:varying_gamma}
\scriptsize
\setlength{\tabcolsep}{3pt}
\renewcommand{\arraystretch}{0.9}
\resizebox{\linewidth}{!}{
\begin{tabular}{lllllllll}
\toprule
Parameters & AAV (0.500) & AMIE (0.224) & E4B (7.743) & GFP (3.572) & LGK (0.020) & Pab1 (0.84) & TEM (1.229) & UBE2I (2.978) \\
\midrule

$\gamma_1=0.1$, $\gamma_2=0.1$          & 0.718 $\pm$ 0.028 & \underline{0.260$\pm$0.005} & 8.139 $\pm$ 0.028 & 3.615 $\pm$ 0.001 & \underline{0.044 $\pm$ 0.001} & \underline{1.738 $\pm$ 0.298} & \textbf{1.233 $\pm$ 0.001} & \underline{2.998 $\pm$ 0.002} \\

$\gamma_1=1.0$, $\gamma_2=0.1$          & 0.725 $\pm$ 0.034 & 0.259 $\pm$ 0.002 & \underline{8.168 $\pm$ 0.032} & \underline{3.618 $\pm$ 0.002} & 0.042 $\pm$ 0.000 & 1.678 $\pm$ 0.293 & \textbf{1.233 $\pm$ 0.002} & \textbf{2.999 $\pm$ 0.002} \\

$\gamma_1=1.0$, $\gamma_2=1.0$          & \underline{0.732 $\pm$ 0.021} & \textbf{0.262 $\pm$ 0.005} & 8.123 $\pm$ 0.040 & \textbf{3.619 $\pm$ 0.001} & 0.042 $\pm$ 0.002 & 1.623 $\pm$ 0.358 & \underline{1.232 $\pm$ 0.001} & \underline{2.998 $\pm$ 0.002} \\

\midrule

$\gamma_1=0.1$, $\gamma_2=1.0$ (SILO)   & \textbf{0.747 $\pm$ 0.014} & \textbf{0.262 $\pm$ 0.004} & \textbf{8.169 $\pm$ 0.052} & \textbf{3.619 $\pm$ 0.002} & \textbf{0.045 $\pm$ 0.003} & \textbf{1.894 $\pm$ 0.348} & \textbf{1.233 $\pm$ 0.003} & 2.997 $\pm$ 0.003 \\
\bottomrule
\end{tabular}
}
\end{table}

\newpage
\subsection {Results reported from ProSpero}
\begin{table}[H]
  \centering
  \caption{Maximum fitness values obtained by all methods on 8 benchmark protein design tasks. Results reported for comparison partners are taken from \cite{kmicikiewicz2025prospero}. \textcolor{SpringGreen}{Green} indicates an improvement in fitness relative to the wild-type $(x_{\mathrm{WT}})$ sequence. \textbf{Bold}: the best per each task. \underline{Underline}: second-best.}
  \label{tab:prospero_max_fitness}
  \footnotesize 
  \setlength{\tabcolsep}{1pt}
  \resizebox{\linewidth}{!}{
  \begin{tabular}{lllllllll}
    \toprule
    \cmidrule(r){1-9}
    Algorithm     & AAV  & AMIE  & E4B  & GFP & LGK  & Pab1 & TEM  & UBE2I \\
    \midrule
    CMA-ES & 0.000$\pm$0.000  & -6.857$\pm$0.257    & -0.429$\pm$0.252  & 1.972$\pm$0.135 & -1.337$\pm$0.021 & 0.553$\pm$0.038 &  0.037$\pm$0.01 &  0.135$\pm$0.178\\
    DyNaPPO     & 0.009$\pm$0.018 & -3.683$\pm$0.575      &  3.924$\pm$0.883  & 3.550$\pm$0.012 & -0.007$\pm$0.015 & 0.783$\pm$0.036 & 0.067$\pm$0.008 & 2.796$\pm$0.059\\
    BO     & \colorbox{SpringGreen}{0.667$\pm$0.024}      & 0.168$\pm$0.056   & 7.442$\pm$0.242 & \colorbox{SpringGreen}{3.584$\pm$ 0.007} & \colorbox{SpringGreen}{0.026$\pm$0.003} & 0.814$\pm$0.081 & 0.682$\pm$0.369 & 2.883$\pm$0.069\\
    PEX     & \colorbox{SpringGreen}{0.665$\pm$0.022}       & \colorbox{SpringGreen}{0.248$\pm$0.007}   & \colorbox{SpringGreen}{8.099$\pm$0.017} & \colorbox{SpringGreen}{3.603$\pm$ 0.003} & \colorbox{SpringGreen}{0.037$\pm$0.001} & \colorbox{SpringGreen}{1.499$\pm$.343} & \colorbox{SpringGreen}{\underline{1.232$\pm$0.000}} & \colorbox{SpringGreen}{2.991$\pm$0.001}\\
    AdaLead     & \colorbox{SpringGreen}{0.683$\pm$0.037}       & \colorbox{SpringGreen}{0.235$\pm$0.002}   & \colorbox{SpringGreen}{8.034$\pm$0.036} & \colorbox{SpringGreen}{3.581$\pm$0.003} & \colorbox{SpringGreen}{0.038$\pm$0.001} & \colorbox{SpringGreen}{\textbf{1.978$\pm$0.188}} & 1.228$\pm$0.002 & \colorbox{SpringGreen}{2.985$\pm$0.002}\\
    CbAS     & 0.000$\pm$0.000       & -8.202$\pm$0.032   & -0.569$\pm$0.092 & 1.858$\pm$0.067 & -1.492$\pm$0.035 & 0.351$\pm$0.043 & 0.019$\pm$0.002 & -0.056$\pm$0.003\\
    GFN-AL     & 0.000$\pm$0.000       & -7.853$\pm$0.270  & 0.160$\pm$0.228 & 2.004$\pm$0.022 & -1.164$\pm$0.118 & 0.507$\pm$0.025 & 0.027$\pm$0.020 & 0.271$\pm$0.443\\
    GFN-AL-$\delta$CS     & \colorbox{SpringGreen}{0.686$\pm$0.021}       & 0.203$\pm$0.005   & \colorbox{SpringGreen}{7.930$\pm$0.055} & \colorbox{SpringGreen}{3.589 $\pm$0.006} & \colorbox{SpringGreen}{0.033$\pm$0.001} & \colorbox{SpringGreen}{1.297$\pm$ 0.337} & 0.701$\pm$0.148 & \colorbox{SpringGreen}{2.984$\pm$0.002}\\
    LatProtRL     & \colorbox{SpringGreen}{0.593$\pm$0.018}       & 0.224$\pm$0.000  & \colorbox{SpringGreen}{7.902$\pm$0.086}  & \colorbox{SpringGreen}{3.590$\pm$0.003} & \colorbox{SpringGreen}{0.020$\pm$0.000} & \colorbox{SpringGreen}{1.122$\pm$0.152} & 1.229 $\pm$0.000 & \colorbox{SpringGreen}{2.983$\pm$0.000}\\
    MLDE     & \colorbox{SpringGreen}{0.555$\pm$0.000}       & \colorbox{SpringGreen}{0.241$\pm$0.003}   & \colorbox{SpringGreen}{7.934$\pm$0.077} & \colorbox{SpringGreen}{3.596$\pm$0.003} & \colorbox{SpringGreen}{0.038$\pm$0.002} & \colorbox{SpringGreen}{0.896$\pm$0.015} & 1.229$\pm$0.000 & \colorbox{SpringGreen}{2.984$\pm$0.003}\\
    ProSpero    & \colorbox{SpringGreen}{\underline{0.720$\pm$0.027}}       & \colorbox{SpringGreen}{\underline{0.246$\pm$0.006}}   & \colorbox{SpringGreen}{\underline{8.114$\pm$0.037}} & \colorbox{SpringGreen}{\underline{3.617$\pm$0.002}} & \colorbox{SpringGreen}{\underline{0.043$\pm$0.002}} & \colorbox{SpringGreen}{1.527$\pm$0.254} & \colorbox{SpringGreen}{1.231$\pm$0.002} & \colorbox{SpringGreen}{\underline{2.993$\pm$0.003}}\\
    \midrule
    SILO & 
    \colorbox{SpringGreen}{\textbf{0.747 $\pm$ 0.014}} & 
    \colorbox{SpringGreen}{\textbf{0.262 $\pm$ 0.004}} & 
    \colorbox{SpringGreen}{\textbf{8.169 $\pm$ 0.052}} & 
    \colorbox{SpringGreen}{\textbf{3.619 $\pm$ 0.002}} & 
    \colorbox{SpringGreen}{\textbf{0.045 $\pm$ 0.003}} & 
    \colorbox{SpringGreen}{\underline{1.894 $\pm$ 0.348}} & 
    \colorbox{SpringGreen}{\textbf{1.233 $\pm$ 0.003}} & 
    \colorbox{SpringGreen}{\textbf{2.997 $\pm$ 0.003}} \\
    \bottomrule
  \end{tabular}
  }
\end{table}

\begin{table}[h]
\centering
\caption{Mean fitness of top-100 sequences generated by each method. Results reported for comparison partners are taken from \cite{kmicikiewicz2025prospero}. Reported values are the mean and standard deviation over 5 runs. \textbf{Bold}: the best per each task. \underline{Underline}: second-best.}
\label{tab:prospero_mean_fitness}
\footnotesize
\setlength{\tabcolsep}{3pt}
\renewcommand{\arraystretch}{0.9}
\resizebox{\linewidth}{!}{
\begin{tabular}{lllllllll}
\toprule
Method & AAV & AMIE & E4B & GFP & LGK & Pab1 & TEM & UBE2I \\
\midrule
CMA-ES        & 0.000 $\pm$ 0.000 & -8.317 $\pm$ 0.029 & -1.009 $\pm$ 0.029 & 1.593 $\pm$ 0.008 & -1.538 $\pm$ 0.008 & 0.232 $\pm$ 0.012 & 0.013 $\pm$ 0.000 & -0.072 $\pm$ 0.004 \\
DynaPPO       & 0.000 $\pm$ 0.000 & -6.493 $\pm$ 0.155 & 0.574 $\pm$ 0.148 & 2.064 $\pm$ 0.068 & -1.020 $\pm$ 0.045 & 0.481 $\pm$ 0.013 & 0.027 $\pm$ 0.002 & 1.600 $\pm$ 0.101 \\
BO            & \colorbox{SpringGreen}{0.618 $\pm$ 0.010} & -0.849 $\pm$ 0.474 & 5.909 $\pm$ 0.785 & 3.538 $\pm$ 0.036 & -0.017 $\pm$ 0.020 & 0.510 $\pm$ 0.047 & 0.606 $\pm$ 0.352 & 2.695 $\pm$ 0.148 \\

PEX           & \colorbox{SpringGreen}{0.620 $\pm$ 0.017} & \colorbox{SpringGreen}{0.238 $\pm$ 0.004} & \colorbox{SpringGreen}{7.948 $\pm$ 0.046} & \colorbox{SpringGreen}{3.597 $\pm$ 0.003} & \colorbox{SpringGreen}{0.033 $\pm$ 0.001} & \colorbox{SpringGreen}{1.307 $\pm$ 0.258} & \underline{1.227 $\pm$ 0.002} & \colorbox{SpringGreen}{2.987 $\pm$ 0.001} \\

AdaLead       & \colorbox{SpringGreen}{0.644 $\pm$ 0.031} & \colorbox{SpringGreen}{0.229 $\pm$ 0.001} & \colorbox{SpringGreen}{7.846 $\pm$ 0.040} & 3.563 $\pm$ 0.007 & \colorbox{SpringGreen}{0.037 $\pm$ 0.001} & \colorbox{SpringGreen}{\underline{1.836 $\pm$ 0.266}} & 1.201 $\pm$ 0.002 & 2.976 $\pm$ 0.003 \\

CbAS          & 0.000 $\pm$ 0.000 & -8.361 $\pm$ 0.025 & -0.820 $\pm$ 0.068 & 1.666 $\pm$ 0.021 & -1.659 $\pm$ 0.023 & 0.162 $\pm$ 0.082 & 0.010 $\pm$ 0.001 & -0.072 $\pm$ 0.003 \\
GFN-AL        & 0.000 $\pm$ 0.000 & -8.268 $\pm$ 0.010 & -0.415 $\pm$ 0.091 & 1.776 $\pm$ 0.009 & -1.345 $\pm$ 0.037 & 0.276 $\pm$ 0.036 & 0.015 $\pm$ 0.001 & 0.172 $\pm$ 0.396 \\

GFN-AL-$\delta$CS & \colorbox{SpringGreen}{0.648 $\pm$ 0.020} & -0.244 $\pm$ 0.137 & 7.653 $\pm$ 0.136 & 3.569 $\pm$ 0.009 & \colorbox{SpringGreen}{0.024 $\pm$ 0.004} & \colorbox{SpringGreen}{1.070 $\pm$ 0.113} & 0.192 $\pm$ 0.027 & 2.968 $\pm$ 0.006 \\

LatProtRL     & \colorbox{SpringGreen}{0.563 $\pm$ 0.009} & 0.217 $\pm$ 0.001 & 7.562 $\pm$ 0.060 & \colorbox{SpringGreen}{3.582 $\pm$ 0.003} & 0.019 $\pm$ 0.000 & \colorbox{SpringGreen}{0.888 $\pm$ 0.072} & \underline{1.222 $\pm$ 0.000 }& 2.975 $\pm$ 0.001 \\

MLDE & \colorbox{SpringGreen}{0.555 $\pm$ 0.000} & \colorbox{SpringGreen}{0.231 $\pm$ 0.004} & \colorbox{SpringGreen}{7.843 $\pm$ 0.122} & \colorbox{SpringGreen}{3.591 $\pm$ 0.003} & \colorbox{SpringGreen}{0.036 $\pm$ 0.002} & \colorbox{SpringGreen}{0.877 $\pm$ 0.024} & 1.131 $\pm$ 0.021 & 2.975 $\pm$ 0.005 \\

ProSpero  & \colorbox{SpringGreen}{\underline{0.679 $\pm$ 0.025}} & \colorbox{SpringGreen}{\underline{0.236 $\pm$ 0.007}} & \colorbox{SpringGreen}{\underline{8.017 $\pm$ 0.054}} & \colorbox{SpringGreen}{\underline{3.613 $\pm$ 0.002}} & \colorbox{SpringGreen}{\underline{0.040 $\pm$ 0.002}} & \colorbox{SpringGreen}{1.401 $\pm$ 0.202} & 1.176 $\pm$ 0.029 & \colorbox{SpringGreen}{\underline{2.987 $\pm$ 0.003}} \\
\midrule
SILO & \colorbox{SpringGreen}{\textbf{0.727 $\pm$ 0.014}} & \colorbox{SpringGreen}{\textbf{0.259 $\pm$ 0.005}} & \colorbox{SpringGreen}{\textbf{8.148 $\pm$ 0.058}} & 
\colorbox{SpringGreen}{\textbf{3.617 $\pm$ 0.001}} & \colorbox{SpringGreen}{\textbf{0.044 $\pm$ 0.002}} & \colorbox{SpringGreen}{\textbf{1.856 $\pm$ 0.355}} & \colorbox{SpringGreen}{\textbf{1.232 $\pm$ 0.001}} & \colorbox{SpringGreen}{\textbf{2.996 $\pm$ 0.002}}\\

\bottomrule
\end{tabular}
}
\end{table}

\clearpage
\input{checklist.tex}

\end{document}

%% file: checklist.tex
\section*{NeurIPS Paper Checklist}
\begin{enumerate}

\item {\bf Claims}
    \item[] Question: Do the main claims made in the abstract and introduction accurately reflect the paper's contributions and scope?
    \item[] Answer: \answerYes{} 
    \item[] Justification: The abstract and introduction accurately reflect SILO's methodological contribution as a trajectory-level protein sequence optimization approach combining stochastic beam search without replacement (SBS), proxy-guided candidate selection complemented by an auxiliary alanine-scan-inspired fitness score (AFS), and iterative policy adaptation through self-improvement imitation learning (SIL). The empirical claims are supported by evaluations across eight protein fitness landscapes in subsection \ref{ssec:sequence_optimization_exp}, including paired statistical comparisons that identify significant improvements on two landscapes for maximum fitness and four for top-100 mean fitness after Holm correction. Additional low-data and noisy-proxy experiments in subsections \ref{ssec:low_data_regime} \ref{ssec:noisy_proxy_setting} and assess performance under selected challenging conditions, without claiming general robustness or convergence beyond the evaluated oracle budgets.
    

\item {\bf Limitations}
    \item[] Question: Does the paper discuss the limitations of the work performed by the authors?
    \item[] Answer: \answerYes{} 
    \item[] Justification: The limitations of our work are mentioned in subsection \ref{ssec:limitations}.

\item {\bf Theory assumptions and proofs}
    \item[] Question: For each theoretical result, does the paper provide the full set of assumptions and a complete (and correct) proof?
    \item[] Answer: \answerNA{} 
    \item[] Justification: We do not include any theoretical results or proofs in this work. 

    \item {\bf Experimental result reproducibility}
    \item[] Question: Does the paper fully disclose all the information needed to reproduce the main experimental results of the paper to the extent that it affects the main claims and/or conclusions of the paper (regardless of whether the code and data are provided or not)?
    \item[] Answer: \answerYes{}
    \item[] Justification: We provide a detailed explanation regarding our action space, policy architecture, and learning algorithm in section \ref{sec:proposed_framework}. Additionally, we also provide code to reproduce the results of experiments in subsections \ref{ssec:sequence_optimization_exp}, \ref{ssec:low_data_regime}, and \ref{ssec:noisy_proxy_setting} under the github repository: \url{https://github.com/grimmlab/SILO.git}. 

\item {\bf Open access to data and code}
    \item[] Question: Does the paper provide open access to the data and code, with sufficient instructions to faithfully reproduce the main experimental results, as described in supplemental material?
    \item[] Answer: \answerYes{} 
    \item[] Justification: We provide code to reproduce the results of experiments in subsections \ref{ssec:sequence_optimization_exp}, \ref{ssec:low_data_regime}, and \ref{ssec:noisy_proxy_setting} and the datasets used to train the proxies under the github repository: \url{https://github.com/grimmlab/SILO.git}. Moreover, details regarding datasets and oracles used, proxy training, and implementation of baselines is discussed. 

\item {\bf Experimental setting/details}
    \item[] Question: Does the paper specify all the training and test details (e.g., data splits, hyperparameters, how they were chosen, type of optimizer) necessary to understand the results?
    \item[] Answer: \answerYes{} 
    \item[] Justification: The setup of our experiments is discussed in subsections \ref{ssec:sequence_optimization_exp}, \ref{ssec:low_data_regime}, and \ref{ssec:noisy_proxy_setting}. We further provide information about the datasets and oracles used, proxy training, and implementation of baselines in section \ref{sec:supplementary_experiments}. 

\item {\bf Experiment statistical significance}
    \item[] Question: Does the paper report error bars suitably and correctly defined or other appropriate information about the statistical significance of the experiments?
    \item[] Answer: \answerYes{} 
    \item[] Justification: We report mean and standard deviation over five independent runs for all experiments in section \ref{sec:experiments}. Results are presented as mean $\pm$ standard deviation in tables, and as shaded regions in figures to show variability across runs.

\item {\bf Experiments compute resources}
    \item[] Question: For each experiment, does the paper provide sufficient information on the computer resources (type of compute workers, memory, time of execution) needed to reproduce the experiments?
    \item[] Answer: \answerYes{} 
    \item[] Justification: We discuss the details about compute workers and execution time in subsection \ref{ssec:hardware}. 
    
\item {\bf Code of ethics}
    \item[] Question: Does the research conducted in the paper conform, in every respect, with the NeurIPS Code of Ethics \url{https://neurips.cc/public/EthicsGuidelines}?
    \item[] Answer: \answerYes{} 
    \item[] Justification: Our work conforms with the NeurIPS Code of Ethics. 

\item {\bf Broader impacts}
    \item[] Question: Does the paper discuss both potential positive societal impacts and negative societal impacts of the work performed?
    \item[] Answer: \answerYes{} 
    \item[] Justification: We discuss the broader impacts of our work in subsection \ref{ssec:broader_impact}. 
    
\item {\bf Safeguards}
    \item[] Question: Does the paper describe safeguards that have been put in place for responsible release of data or models that have a high risk for misuse (e.g., pre-trained language models, image generators, or scraped datasets)?
    \item[] Answer: \answerNA{} 
    \item[] Justification: This work does not involve the release of models or datasets that pose a high risk of misuse.

\item {\bf Licenses for existing assets}
    \item[] Question: Are the creators or original owners of assets (e.g., code, data, models), used in the paper, properly credited and are the license and terms of use explicitly mentioned and properly respected?
    \item[] Answer: \answerYes{} 
    \item[] Justification: All asset owners are properly credited, and licenses explicitly mentioned and
respected.

\item {\bf New assets}
    \item[] Question: Are new assets introduced in the paper well documented and is the documentation provided alongside the assets?
    \item[] Answer: \answerYes{}
    \item[] Justification: We will release documented code implementing our method under the GPL-3.0 license.

\item {\bf Crowdsourcing and research with human subjects}
    \item[] Question: For crowdsourcing experiments and research with human subjects, does the paper include the full text of instructions given to participants and screenshots, if applicable, as well as details about compensation (if any)? 
    \item[] Answer: \answerNA{}
    \item[] Justification: This work does not involve crowdsourcing or research with human subjects.

\item {\bf Institutional review board (IRB) approvals or equivalent for research with human subjects}
    \item[] Question: Does the paper describe potential risks incurred by study participants, whether such risks were disclosed to the subjects, and whether Institutional Review Board (IRB) approvals (or an equivalent approval/review based on the requirements of your country or institution) were obtained?
    \item[] Answer: \answerNA{}
    \item[] Justification: This work does not involve research with human subjects.

\item {\bf Declaration of LLM usage}
    \item[] Question: Does the paper describe the usage of LLMs if it is an important, original, or non-standard component of the core methods in this research? Note that if the LLM is used only for writing, editing, or formatting purposes and does \emph{not} impact the core methodology, scientific rigor, or originality of the research, declaration is not required.
    \item[] Answer: \answerNA{}
    \item[] Justification: This work does not involve LLMs as an important, original, or non-standard
component of the core methodology.

\end{enumerate}